\documentclass{article}

\usepackage{microtype}
\usepackage{graphicx}
\usepackage{booktabs} 

\usepackage{hyperref}

\usepackage{amsmath,amsfonts,bm}

\def\eqref#1{equation~\ref{#1}}

\def\1{\bm{1}}

\DeclareMathAlphabet{\mathsfit}{\encodingdefault}{\sfdefault}{m}{sl}
\SetMathAlphabet{\mathsfit}{bold}{\encodingdefault}{\sfdefault}{bx}{n}

\usepackage{url}
\usepackage{graphicx}
\usepackage{booktabs}       
\usepackage{amsfonts}       
\usepackage{nicefrac}       
\usepackage{microtype}      
\usepackage{xcolor}         
\usepackage{graphicx}
\usepackage{subcaption}
\usepackage{caption}
\usepackage{wrapfig}
\usepackage{algorithm,algorithmic}
\usepackage{multirow}
\usepackage{tcolorbox}
\usepackage{overpic}
\usepackage{makecell}

\usepackage[accepted]{icml2024}

\usepackage{amsmath}
\usepackage{amssymb}

\usepackage{mathtools}
\usepackage{amsthm}
\usepackage{enumitem}

\usepackage[capitalize,noabbrev]{cleveref}

\theoremstyle{plain}
\newtheorem{theorem}{Theorem}[section]
\newtheorem{proposition}[theorem]{Proposition}

\theoremstyle{definition}
\newtheorem{definition}[theorem]{Definition}

\theoremstyle{remark}

\definecolor{eqgreen}{rgb}{0, 0.69, 0.31}

\usepackage[textsize=tiny]{todonotes}

\icmltitlerunning{HarmonyDream: Task Harmonization Inside World Models}

\begin{document}

\twocolumn[

\icmltitle{HarmonyDream: Task Harmonization Inside World Models}



\icmlsetsymbol{equal}{*}

\begin{icmlauthorlist}
\icmlauthor{Haoyu Ma}{equal,bnrist}
\icmlauthor{Jialong Wu}{equal,bnrist}
\icmlauthor{Ningya Feng}{bnrist}
\icmlauthor{Chenjun Xiao}{huawei}
\icmlauthor{Dong Li}{huawei}
\icmlauthor{Jianye Hao}{huawei,tju}
\icmlauthor{Jianmin Wang}{bnrist}
\icmlauthor{Mingsheng Long}{bnrist}
\end{icmlauthorlist}

\icmlaffiliation{bnrist}{School of Software, BNRist, Tsinghua University.}
\icmlaffiliation{huawei}{Huawei Noah's Ark Lab.}
\icmlaffiliation{tju}{College of Intelligence and Computing, Tianjin University.

Haoyu Ma $<$mhy22@mails.tsinghua.edu.cn$>$. Jialong Wu $<$wujialong0229@gmail.com$>$}

\icmlcorrespondingauthor{Mingsheng Long}{mingsheng@tsinghua.edu.cn}

\icmlkeywords{Reinforcement learning, World model}

\vskip 0.3in
]



\printAffiliationsAndNotice{\icmlEqualContribution} 

\begin{abstract}
Model-based reinforcement learning (MBRL) holds the promise of sample-efficient learning by utilizing a world model, which models how the environment works and typically encompasses components for two tasks: observation modeling and reward modeling. In this paper, through a dedicated empirical investigation, we gain a deeper understanding of the role each task plays in world models and uncover the overlooked potential of sample-efficient MBRL by mitigating the domination of either observation or reward modeling. Our key insight is that while prevalent approaches of explicit MBRL attempt to restore abundant details of the environment via observation models, it is difficult due to the environment's complexity and limited model capacity. On the other hand, reward models, while dominating implicit MBRL and adept at learning compact task-centric dynamics, are inadequate for sample-efficient learning without richer learning signals. Motivated by these insights and discoveries, we propose a simple yet effective approach, \textit{HarmonyDream}, which automatically adjusts loss coefficients to maintain \textit{task harmonization}, i.e.~a dynamic equilibrium between the two tasks in world model learning. Our experiments show that the base MBRL method equipped with HarmonyDream gains 10\%$-$69\% absolute performance boosts on visual robotic tasks and sets a new state-of-the-art result on the Atari 100K benchmark. Code is available at \url{https://github.com/thuml/HarmonyDream}.
\end{abstract}

\section{Introduction}

Learning efficiently to operate in environments with complex observations requires generalizing from past experiences. 
Model-based reinforcement learning (MBRL, \citet{sutton1990integrated}) utilizing world models \citep{ha2018recurrent, lecun2022path} offers a promising approach. In MBRL, the agent learns behaviors by simulating trajectories based on world model predictions. These imaginary rollouts can reduce the need for real-environment interactions, thus improving the sample efficiency of model-based agents.

Concretely, world models are designed to learn and predict two key components of dynamics (formally defined in Sec.~\ref{sec:multitask}): how the environment transits and is observed (i.e. the \textit{observation modeling} task) and how the task has been progressed (i.e. the \textit{reward modeling} task) \citep{kaiser2019model, hafner2019dream, hafner2020mastering}. While observation transitions and rewards in low-dimensional spaces can be classically learned by separate models, for environments with high-dimensional and partial observations, it is favorable for world models to learn both tasks from a shared representation, a form of \textit{multi-task learning}\footnote{Here we refer to intrinsic multi-task learning inside world models rather than multi-task policy learning for different rewards.} \cite{caruana1997multitask}, aiming to improve learning efficiency and generalization performance \citep{jaderberg2016reinforcement, laskin2020curl, yarats2021improving}. However, to best exploit the benefits of multi-task learning, it demands careful designs to weigh the contribution of each task without allowing either one to dominate \citep{misra2016cross, kendall2018multitask}, which naturally leads to the following question:
\begin{center}
\textit{How do MBRL methods properly exploit the intrinsic multi-task benefits within world model learning?}
\end{center}

\begin{figure*}[ht]
    \centering
    \includegraphics[width=0.9\textwidth]{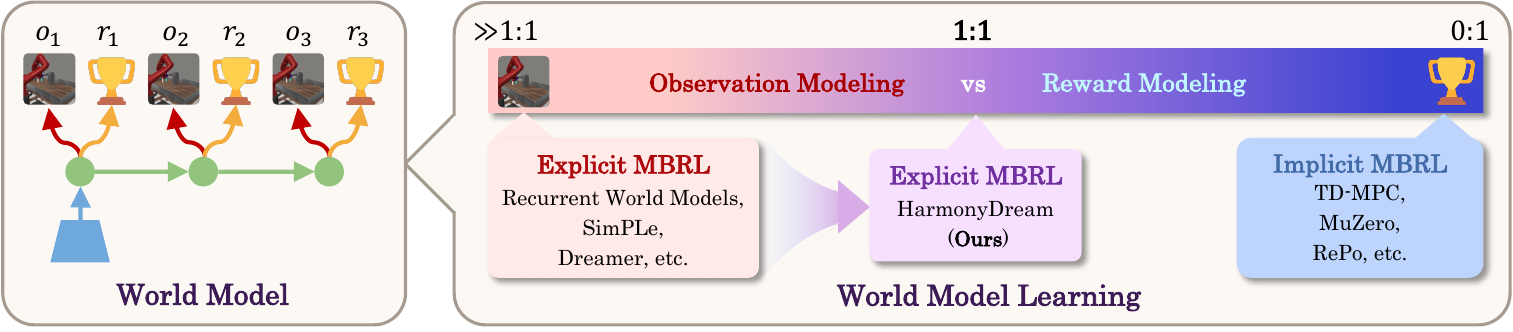}
    \vspace{-3pt}
    \caption{A multi-task view of world models. (\textit{Left}) World models typically consist of components for two tasks: \textcolor[HTML]{C00000}{\textbf{observation modeling}} and \textcolor[HTML]{F3AC3E}{\textbf{reward modeling}}. (\textit{Right}) A spectrum of world model learning in MBRL. Explicit MBRL learns models dominated by observation modeling, while implicit MBRL relies solely on reward modeling. In the spirit of implicit MBRL, our proposed HarmonyDream enables explicit MBRL to maintain a dynamic equilibrium between them to unleash the multi-task benefits of world model learning, thus boosting the sample efficiency of MBRL.}
    \label{fig:spectrum}
    \vspace{-3pt}
\end{figure*}

In this work, we take a unified multi-task view to revisit world model learning in MBRL literature \citep{moerland2023survey}: Prevalent \textit{explicit} MBRL approaches \citep{kaiser2019model, hafner2020mastering, seo2022reinforcement}, which is also our primary focus, aim to learn an exact duplicate of the environment by predicting each element (e.g., observations, rewards, and terminal signals), which gives the agent access to accurately learned transitions. However, learning to predict future observations can be difficult and inefficient since it encourages the world model to capture everything in the environment, including task-irrelevant nuances \citep{okada2021dreaming, deng2022dreamerpro}. Consequently, world model learning in explicit MBRL is typically dominated by observation modeling to capture complex observations and their associated dynamics but still suffers from model inaccuracies and compounding errors. 
This can be overcome by the spirit of \textit{implicit} MBRL, which learns task-centric world models solely from reward modeling \citep{oh2017value, schrittwieser2020mastering, hansen2022temporal} to realize the value equivalence principle, i.e., the predicted rewards along a trajectory of the world model matches that of the real environment \citep{grimm2020value}. This approach builds world models directly useful for MBRL to identify the optimal policy or value, and tends to perform better in tasks where the complete dynamics related to observations are too complicated to be perfectly modeled. Nevertheless, as the reward signals in RL are known to be sparser than signals in self-supervised learning, potentially leading to representation learning challenges, it is more practical to incorporate auxiliary tasks that provide richer learning signals beyond rewards \citep{jaderberg2016reinforcement, anand2021procedural}.

To support the above insights, we conduct a dedicated empirical investigation and reveal surprising deficiencies in sample efficiency within the default practice of a state-of-the-art model-based method (Dreamer, \citet{hafner2019dream,hafner2020mastering, hafner2023mastering}). Notably, \textit{increasing the coefficient of reward loss in world model learning leads to dramatically boosted sample efficiency} (see Sec.~\ref{sec:observaions}). Our analysis identifies the root cause as the domination of observation models in explicit world model learning: due to an overload of redundant observation signals, the model may establish spurious correlations in observations without realizing incorrect reward predictions, which ultimately hinders the learning process of the agent. On the other hand, a pure implicit version of Dreamer, which learns world models solely exploiting reward modeling, is also proven to be inefficient. In summary, domination of either task cannot properly exploit the multi-task benefits within world model learning.

As shown in Fig.~\ref{fig:spectrum}, we propose to address the problem with \textbf{HarmonyDream}, a simple approach for explicit world model learning that exploits the advantages of both sides. By automatically adjusting loss coefficients through lightweight harmonizers, HarmonyDream seeks task harmonization inside world models, i.e., it maintains a dynamic equilibrium between reward and observation modeling during world model learning. We evaluate our approach on various challenging visual control domains, including Meta-world \citep{yu2020meta}, RLBench \citep{james2020rlbench}, distracted DMC variants \citep{grigsby2020measuring, zhang2018natural}, the Atari 100K benchmark \citep{kaiser2019model}, and a challenging task from Minecraft \citep{fan2022minedojo}, demonstrating consistent improvements in sample efficiency and generality to different base MBRL approaches \citep{deng2022dreamerpro}.

The main contributions of this work are three-fold:
\vspace{-5pt}
\begin{itemize}
    \item To the best of our knowledge, our work, for the first time, systematically identifies the multi-task essence of world models and analyzes the deficiencies caused by the domination of a particular task, which is unexpectedly overlooked by most previous works.
    \vspace{-2pt}
    \item We propose HarmonyDream, a simple yet effective world model learning approach to mitigate the domination of either observation or reward modeling, without the need for exhaustive hyperparameter tuning.
    \vspace{-2pt}
    \item Our experiments show that HarmonyDream improves Dreamer with 10\%$-$69\% higher success rates or episode returns (up to 90\% more success on Meta-world Assembly) in visual robotic tasks. Moreover, our method reaches a new state of the art, 136.5\% mean human performance, on the Atari 100K benchmark.
\end{itemize}

\section{A Multi-task Analysis in World Models}

\begin{figure*}[tbp]
    \centering
    \begin{subfigure}[t]{0.45\textwidth}
        \centering
        \includegraphics[width=\textwidth]{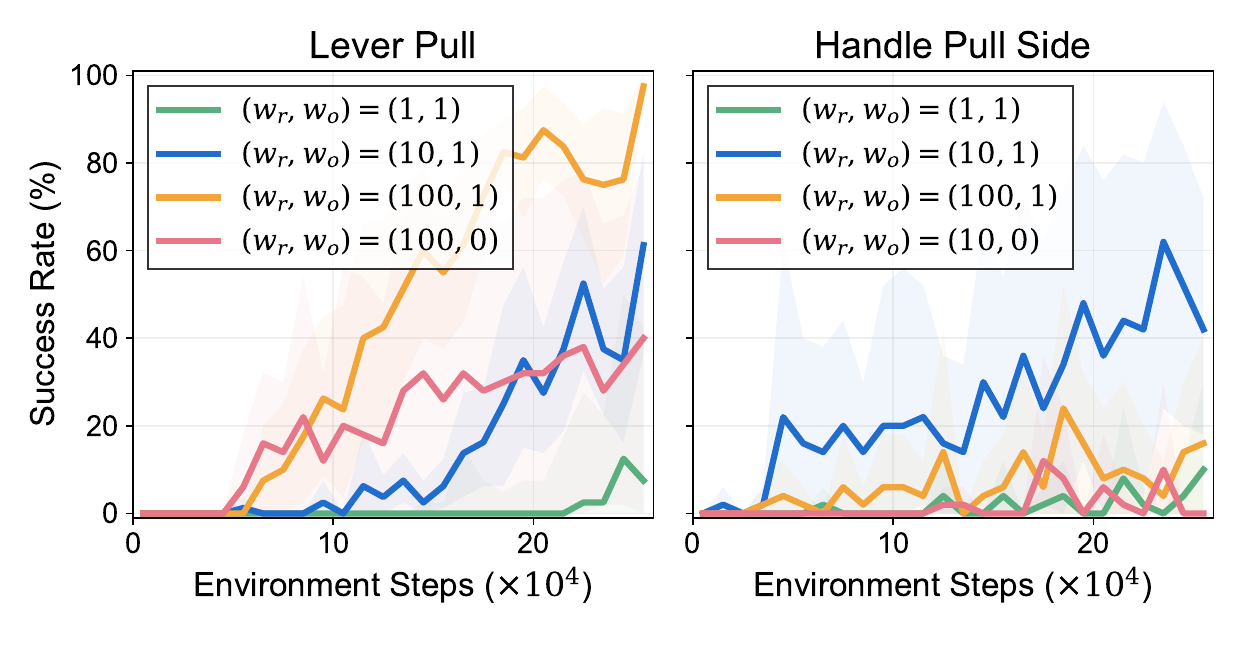}
        {
        \vspace{-18pt}
        \caption{Learning curves of different loss coefficients}
        \label{fig:effect_coeff_metaworld}}
    \end{subfigure}
    ~
    \hspace{-10pt}
    \begin{subfigure}[t]{0.26\textwidth}
        \centering
        \raisebox{0.087\height}{
        \includegraphics[width=\textwidth]{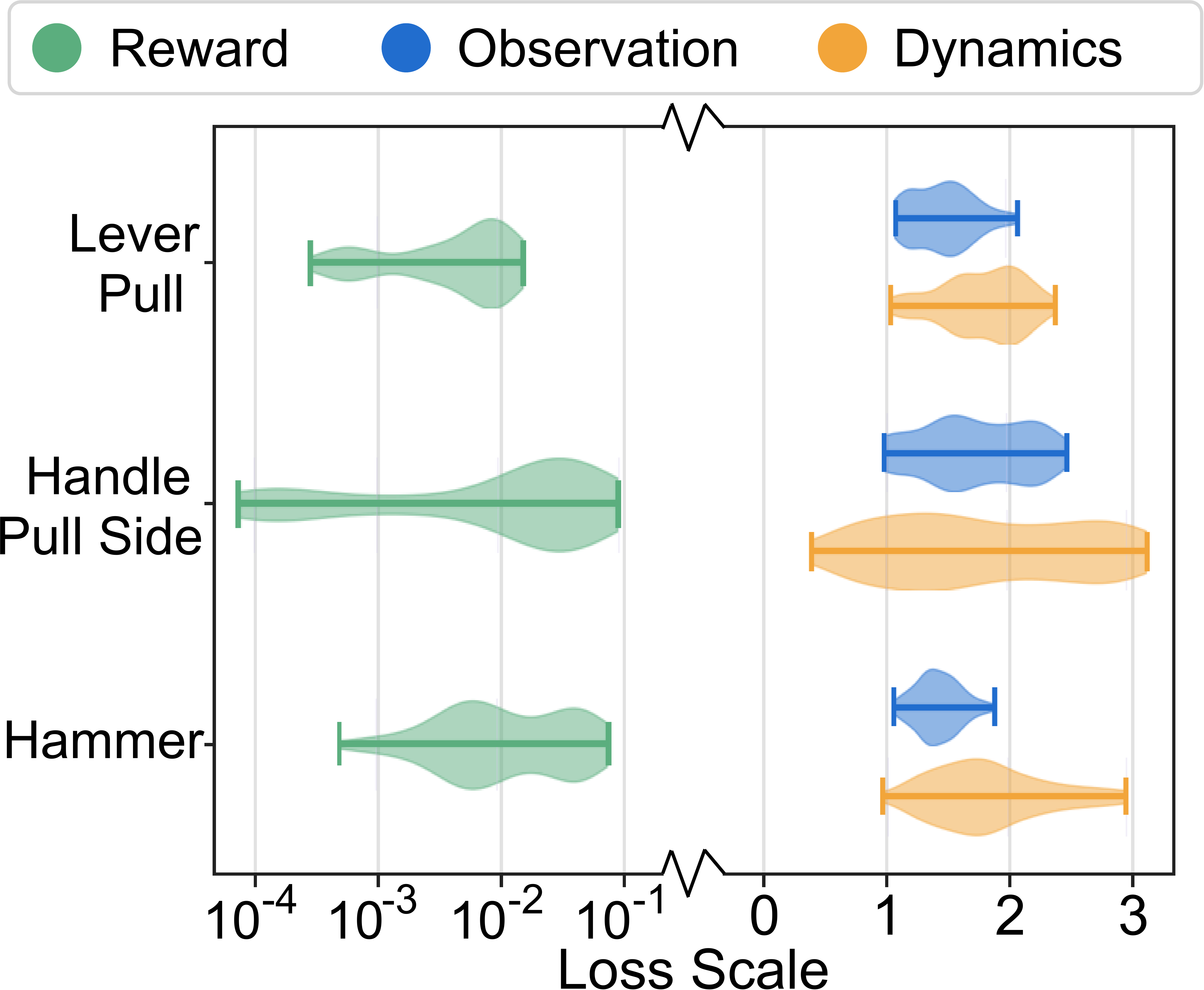}}
        {
        \vspace{-18pt}
        \caption{Loss scales}
        \label{fig:loss_scales_difference}}
    \end{subfigure}
    ~
    \hspace{-10pt}
    \begin{subfigure}[t]{0.26\textwidth}
        \centering
        \raisebox{0.05\height}{
        \includegraphics[width=\textwidth]{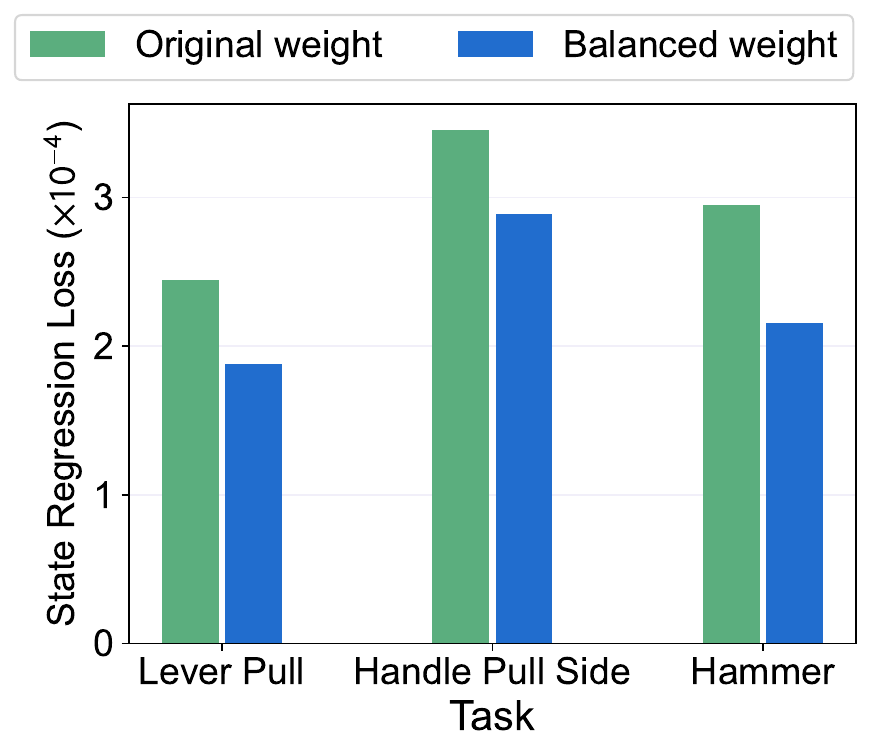}}
        {
        \vspace{-18pt}
        \caption{State regression}
        \label{fig:representation_difference}}
        
    \end{subfigure}
    {
    \vspace{-5pt}
    \caption{Analysis experiments revealing the (b) imbalanced nature of world model learning and potential multi-task benefits yet to be properly exploited. Simply adjusting the coefficient of reward loss leads to (a) dramatically boosted sample efficiency of DreamerV2 agents and (c) better representations with lower environment state regression errors.}
    \label{fig:analysis_fig}}
\end{figure*}

In this paper, we focus on vision-based RL tasks, formulated as partially observable Markov decision processes (POMDP). A POMDP is defined as a tuple $(\mathcal{O}, \mathcal{A}, p, r, \gamma)$, where actions $a_t\sim \pi(a_t\,|\,o_{\leq t}, a_{<t})$ generated by the agent receive high-dimensional observations $o_t \sim p\left(o_t\,|\, o_{<t}, a_{<t}\right)$ and scalar rewards $r_t=r(o_{\leq t}, a_{<t})$ generated by the unknown transition dynamics $p$ and reward function $r$. The goal of MBRL is to learn an agent that maximizes the $\gamma$-discounted cumulative rewards $\mathbb{E}_{p,\pi}\left[\sum_{t=1}^T \gamma^{t-1} r_t\right]$, leveraging a learned world model which approximates the underlying environment $(p,r)$.

\subsection{Two tasks in World Models}
\label{sec:multitask}

Two key tasks can be formally identified in world models, namely observation modeling and reward modeling.
\begin{definition}
The \textit{observation modeling} task in world models is to predict consequent observations $p(o_{t+1:T}\,|\,o_{1:t}, a_{1:T})$ of a trajectory, given future actions. Similarly, the \textit{reward modeling} task in world models is to predict future rewards $p(r_{t+1:T}\,|\,o_{1:t},a_{1:T})$.
\end{definition}
\vspace{-5pt}

As mentioned before, these two tasks provide a unified view of MBRL: while explicit MBRL learns world models for both observations and rewards to mirror the complete dynamics of the environment, implicit MBRL only learns from reward modeling to capture task-centric dynamics.

\subsection{Overview of World Model Learning}
\label{sec:overview}

We conduct detailed analysis and build our method primarily upon DreamerV2\footnote{When we started this work, DreamerV3 had not been released. A detailed discussion with DreamerV3 is included in later sections.} \citep{hafner2020mastering}, but we also demonstrate the generality of our method to various base MBRL algorithms, including DreamerV3 \citep{hafner2023mastering} and DreamerPro \citep{deng2022dreamerpro} (see Sec.~\ref{sec:generality}).

The world model in Dreamer (left in Fig.~\ref{fig:spectrum}) consists of the following four components:
\vspace{-3pt}
\begin{gather*}
\begin{aligned}
&\text{Representation model:} &&z_t\sim q_\theta(z_{t} \,|\,z_{t-1},a_{t-1}, o_{t}) \\
&\text{Observation model:} &&\hat{o}_t\sim p_\theta(\hat{o}_{t} \,|\,z_{t})\\
\end{aligned}
\end{gather*}
\begin{gather}
\begin{aligned}
&\text{Transition model:} &&\hat{z}_t\sim p_\theta(\hat{z}_{t} \,|\,z_{t-1}, a_{t-1}) \\
&\text{Reward model:} &&\hat{r}_t \sim p_\theta\left(\hat{r}_t  \,|\, z_t\right).
\label{eq:dreamer}
\end{aligned}
\end{gather}
The latent representation $z_t$ is generated by the representation model using the previous latent state $z_{t-1}$, the current action $a_{t-1}$, and the current visual observation $o_t$. The latent prediction $\hat{z_t}$, meanwhile, is generated by the transition model using only the previous state and current action. All model parameters $\theta$ are trained to learn the observations, rewards, and transitions of the environment by jointly minimizing the following objectives:
\begin{align}
    \label{eq:dreamer_loss}
    &\text{Observation loss:} &&\mathcal{L}_o(\theta) = -\log p_{\theta}(o_{t}\,|\,z_{t}) \nonumber \\
    &\text{Reward loss:} &&\mathcal{L}_r(\theta) = -\log p_{\theta}(r_{t}\,|\,z_{t})  \\
    &\text{Dynamics loss:} &&\mathcal{L}_d(\theta) = \text{KL}[q_{\theta}(z_{t}\,|\,z_{t-1},a_{t-1}, o_{t}) \nonumber \\ & && \quad\quad\quad\quad\quad\,\,\Vert\,p_{\theta}(\hat{z}_{t}\,|\,z_{t-1}, a_{t-1}) ], \nonumber
\end{align}
where the dynamics loss simultaneously trains the latent predictions toward the representations, and regularizes the representations to be predictable. In practice, the observation model and reward model typically leverage Gaussian distributions, and both losses take the form of a simple $L_2$ loss between prediction $\hat{o}_t, \hat{r}_t$ and ground truth $o_t, r_t$ respectively, excluding irrelevant constants.

Taking our multi-task view, the observation model and reward model with their associated losses account for the aforementioned two tasks in the world model of Dreamer. However, they do not operate in isolation and instead interact with and regularize each other upon the shared representation and transition model, in pursuit of either complete or task-centric latent dynamics, respectively.

\begin{figure*}[t]
    \centering
    \includegraphics[width=0.95\textwidth]{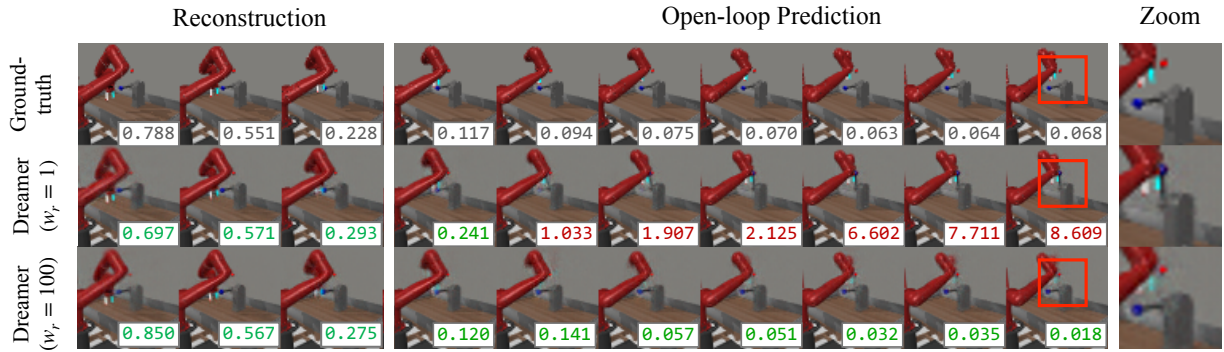}
    \vspace{-5pt}
    \caption{Analysis of world models learned with different reward loss coefficients. Rewards are labeled at the bottom right corner, with predictions marked as \textbf{\color{eqgreen}correct} or \textbf{\color{red}incorrect}. Dominating observation modeling in world models incurs spurious correlations between actions, observations, and rewards, which can be dissolved by properly emphasizing reward modeling.}
    \label{fig:r1_r100_video_prediction_mw}
    \vspace{-4pt}
\end{figure*}

The overall objective can be formulated as follows:
\begin{align}
\label{eq:overall_loss}
\mathcal{L}(\theta) = w_o\mathcal{L}_{o}(\theta)+w_r\mathcal{L}_{r}(\theta)+w_d\mathcal{L}_{d}(\theta).
\end{align}
By default, $w_o$, $w_r$, and $w_d$ are typically set to approximately equal weights (namely, $w_o=w_r=w_d=1.0$) \citep{hafner2019dream,hafner2020mastering,seo2022reinforcement,wu2022daydreamer}, overlooking the potential domination of a particular task. In contrast, we conduct a careful empirical investigation to understand the role each task plays in world models and reveal the deficiency of the default weighting strategy.

\subsection{Dive into World Model Learning}
\label{sec:observaions}

We consider the tasks of \textit{pulling a lever up}, \textit{pulling a handle up sideways}, and \textit{hammering a screw on the wall}, from the Meta-world domain \citep{yu2020meta}, as our testbed to investigate world model learning. The prominent improvements of the derived approach in our benchmark experiments (see Sec.~\ref{sec:exp}) prove that our discoveries can be generalized to various domains and tasks.

First of all, we experiment with simply adjusting the coefficient of the reward loss in Eq. (\ref{eq:overall_loss}). Results in Fig.~\ref{fig:effect_coeff_metaworld} reveal a surprising fact that by simply tuning the reward loss weights ($w_r \in \{1, 10, 100\}$), the agent can achieve considerable improvements in terms of sample efficiency.

\begin{tcolorbox}[colback=blue!2!white,leftrule=2.5mm,size=title]
\textbf{Finding 1.} \textit{Leveraging the reward loss by adjusting its coefficient in world model learning has a great impact on the sample efficiency of model-based agents.}
\end{tcolorbox}

One obvious reason for this is that the reward loss only accounts for a tiny proportion of the learning signals, actually a single scalar $r_t$. As shown in Fig.~\ref{fig:loss_scales_difference}, the scale of $\mathcal{L}_r$ is two orders of magnitude smaller than that of $\mathcal{L}_o$, which usually aggregates $H\times W\times C$ dimensions: $\log p_\theta\left(o_t  \,|\, z_t\right)=\sum_{h,w,c} \log p_{\theta}(o_{t}^{(h,w,c)} \,|\,z_{t})$. As discussed before, reward modeling is crucial for extracting task-relevant representations and driving behavior learning of the agents. Dominated by observation modeling, the world model fails to learn a task-centric latent space and predict accurate rewards, which hinders the agent learning process.

We then explore further to demonstrate how the observation modeling task dominating world models can specifically hurt behavior learning. To isolate distracting factors, we consider an offline setting \citep{levine2020offline}. Concretely, we use a fixed replay buffer on the task of Lever Pull and offline train DreamerV2 agents with different reward loss coefficients on it (see details in Appendix \ref{app:qualitative}). In Fig.~\ref{fig:r1_r100_video_prediction_mw}, we showcase a trajectory where the default Dreamer agent ($w_r=1$) fails to lift the lever. It is evident that it learns a spurious correlation \citep{geirhos2020shortcut} between the actions of the robot and that of the lever and predicts inaccurate transitions and rewards, which misleads the agents to unfavorable behaviors. Properly balancing the reward loss ($w_r=100$) can emphasize task-relevant information, such as whether the lever is actually lifted, to correct hallucinations by world models. Quantitative analysis in Fig.~\ref{fig:representation_difference} measuring the ability of world models' representations to predict the ground truth states also suggests emphasizing reward modeling learns better task-centric representations.

\begin{tcolorbox}[colback=blue!2!white,leftrule=2.5mm,size=title]
\textbf{Finding 2.} \textit{Observation modeling as a dominating task can result in world models establishing spurious correlations without realizing incorrect reward predictions.}
\end{tcolorbox}

Although we have shown above that exploiting reward modeling can bring benefits to world models and MBRL, learning world models depending solely on scarce reward signals, as implicit MBRL, has limited capability to learn meaningful representations, and thus can encounter optimization challenges and hinder sample-efficient learning \citep{yarats2021improving}. Our experiment results in Fig.~\ref{fig:effect_coeff_metaworld} show that a pure implicit version of DreamerV2 without the observation loss ($w_o=0$) produces inferior results with a high variance.

\begin{figure*}[!h]
    \centering
    \begin{minipage}[t]{0.63\textwidth}
        \centering
        \includegraphics[width=\textwidth]{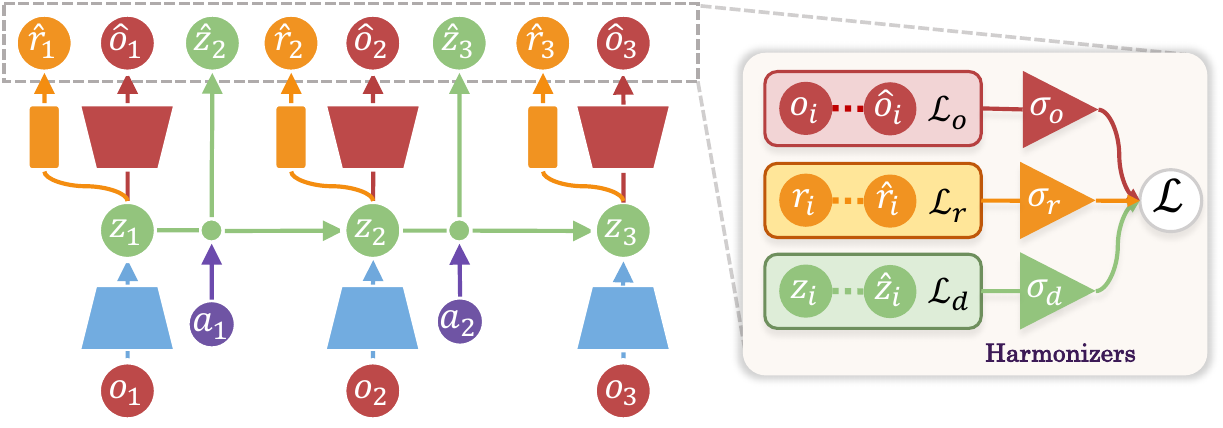}
    \end{minipage}
    ~
    \hspace{10pt}
    \begin{minipage}[t]{0.27\textwidth}
        \centering
        \includegraphics[width=\textwidth]{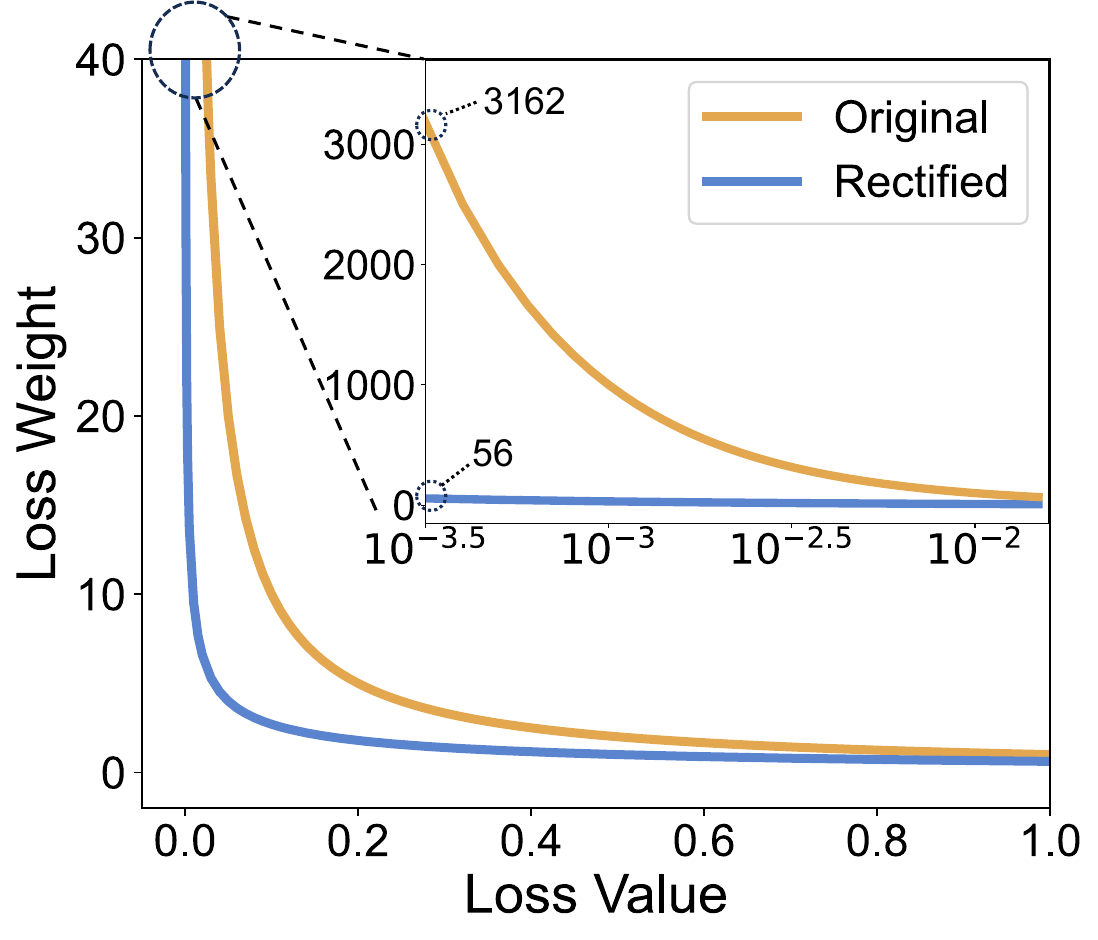}
    \end{minipage}
    \vspace{-3pt}
    \caption{Overview of HarmonyDream. (\textit{Left}) Built upon Dreamer, we introduce lightweight harmonizers to maintain a dynamic equilibrium between tasks. (\textit{Right}) Comparison between the original harmonious loss (Eq.~(\ref{eq:harmony_loss})) and the rectified one (Eq.~(\ref{eq:harmony_loss_with_base})). The latter prevents an extremely large loss weight.}
    \label{fig:HarmonyDream}
\end{figure*}

\begin{tcolorbox}[colback=blue!2!white,leftrule=2.5mm,size=title]
\textbf{Finding 3.} \textit{Learning signal of world models from rewards alone without observations is inadequate for sample-efficient model-based learning.}
\end{tcolorbox}

\vspace{-5pt}
\paragraph{Discussion.} We are not the first to adjust loss coefficients in world model learning, but we dedicatedly investigate this. Here we discuss the differences between our findings and previous literature. Our Finding 1 coincides with high reward loss weights manually tuned (typically 100 or 1000) in decoder-free model-based RL \citep{nguyen2021temporal, deng2022dreamerpro}. Our analysis differs from theirs in two significant ways: 1) We focus on a decoder-based world model, where the observations are learned from explicit reconstructions. 2) We discovered that emphasizing reward modeling is also beneficial for visually simple tasks (e.g. Meta-world tasks), in addition to visually demanding tasks with noisy backgrounds. Our Finding 3 is similar to the reward-only ablation in Dreamer \citep{hafner2019dream}, but we prove that even if given higher loss weights, learning a world model purely from rewards is less sample-efficient than properly exploiting both observation and reward modeling.

\section{HarmonyDream}
\label{sec:method}

In light of the discoveries and insights, we propose a simple yet effective method, HarmonyDream, as the first step towards exploiting the multi-task essence of explicit world model learning. Instead of task domination, we aim to dynamically maintain a harmonious interaction between the two tasks in world models: while observation modeling facilitates representation learning and prevents information loss, reward modeling enhances task-centric representations to inform behavior learning of the agents.

HarmonyDream mitigates the potential domination of a particular task in world models by introducing lightweight harmonizers, as shown in Fig.~\ref{fig:HarmonyDream}. Specifically, to maintain a dynamic equilibrium and avoid task domination, losses associated with different tasks are scaled to the same constant. A straightforward but suboptimal way is to set each loss weight to the reciprocal of the corresponding loss, i.e., $w_i={\rm{sg}}(\frac{1}{\mathcal{L}_i}), i\in \{o,r,d\}$, where $\rm{sg}$ is a stop gradient function. Technically, as the loss is only calculated from a mini-batch of data and fluctuates throughout training, these weights are sensitive to outlier values and thus may further aggravate training instability. 
Instead, we adopt a variational method to learn the weights of different losses by the following \textit{harmonious loss} for world model learning:
\textcolor{white}{}
\vspace{-5pt}
\begin{gather}
\begin{aligned}
    \label{eq:harmony_loss}
    \mathcal{L}(\theta, \sigma_{o}, \sigma_{r}, \sigma_{d}) &= \sum_{i\in \{o,r,d\}}\mathcal{H}(\mathcal{L}_i(\theta), \sigma_i) \\
 &= \sum_{i\in \{o,r,d\}}\frac{1}{{\sigma_i}}\mathcal{L}_{i}(\theta) +\log{\sigma_i}.
\end{aligned} 
\end{gather}
The variational formulation $\mathcal{H}(\mathcal{L}_i(\theta), \sigma_i)=\sigma_i^{-1}\mathcal{L}_{i}(\theta) +\log{\sigma_i}$ serves as \textit{harmonizers} to dynamically but smoothly rescale different losses, where the weight $\sigma_i^{-1}$ with a learnable parameter $\sigma_i>0$ approximates a ``global" reciprocal of the loss scale, as stated in the following proposition:
\begin{proposition}
    The optimal solution $\sigma^*$ that minimizes the expected loss $\mathbb{E}[\mathcal{H}(\mathcal{L}, \sigma)]$, or equivalently $\nabla_{\sigma}\mathbb{E}[\mathcal{H}(\mathcal{L}, \sigma)] =0$, is $\sigma^* = \mathbb{E}[\mathcal{L}]$. In other words, the harmonized loss scale is $\mathbb{E}[\mathcal{L}/\sigma^*]=1$.
    \label{prop:harmony}
\end{proposition}

In practice, $\sigma_i$ is parameterized as $\sigma_i=\exp(s_i)>0$, in order to optimize parameters $s_i$ free of sign constraint. More essentially, we propose a rectification on Eq.~(\ref{eq:harmony_loss}), since a loss $\mathcal{L}$ with small values, such as the reward loss, can lead to extremely large coefficient $1/\sigma\approx \mathcal{L}^{-1} \gg 1$, which potentially hurt training stability. Specifically, we simply add a constant in regularization terms:
\begin{gather}
\begin{aligned}
    \label{eq:harmony_loss_with_base}
    \mathcal{L}(\theta, \sigma_{o}, \sigma_{r}, \sigma_{d}) &= \sum_{i\in \{o,r,d\}}\hat{\mathcal{H}}(\mathcal{L}_i(\theta), \sigma_i) \\
 &= \sum_{i\in \{o,r,d\}}\frac{1}{{\sigma_i}}\mathcal{L}_{i}(\theta) +\log{\left(1+\sigma_i\right)}.
\end{aligned}
\end{gather}
The harmonized loss scale by the \textit{rectified harmonious loss} is equal to $\frac{2}{1+\sqrt{1+4/\mathbb{E}[\mathcal{L}]}} < 1$ (derived in Appendix \ref{app:derivations}). We illustrate the corresponding loss weights learned with different loss scales in the right of Fig.~\ref{fig:HarmonyDream}, showing that the rectified loss effectively mitigates extremely large weights.

\begin{figure*}[ht]
    \centering
    \begin{subfigure}[t]{0.22\textwidth}
        \centering
        \includegraphics[width=\textwidth]{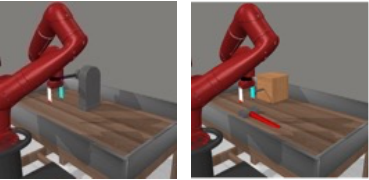}
        {
        \vspace{-12pt}
        \caption{Meta-world}
        }
    \end{subfigure}
    ~
    \begin{subfigure}[t]{0.22\textwidth}
        \centering
        \includegraphics[width=\textwidth]{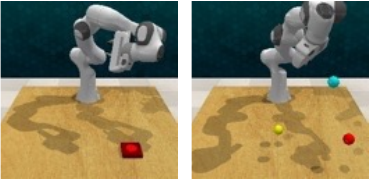}
        {
        \vspace{-12pt}
        \caption{RLBench}
        }
    \end{subfigure}
    ~
    \begin{subfigure}[t]{0.22\textwidth}
        \centering
        \includegraphics[width=\textwidth]{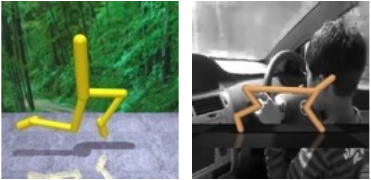}
        {
        \vspace{-12pt}
        \caption{Distracted DMC variants}
        }
    \end{subfigure}
    ~
    ~
    \begin{subfigure}[t]{0.107\textwidth}
        \centering
        \includegraphics[width=\textwidth]{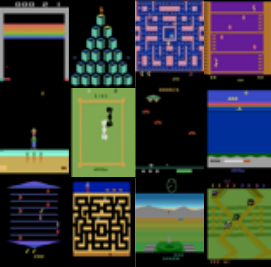}
        {
        \vspace{-12pt}
        \caption{Atari}
        }
    \end{subfigure}
    ~
    \begin{subfigure}[t]{0.106\textwidth}
        \centering
        \includegraphics[width=\textwidth]{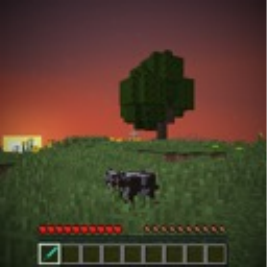}
        {
        \vspace{-12pt}
        \caption{Minecraft}
        }
    \end{subfigure}
    {
    \vspace{-5pt}
    \caption{Visual control domains for evaluation, including robotic manipulation (a-b), distracted locomotion (c), and video games (d-e).}
    \label{fig:example_obs_single}}
\end{figure*}

\vspace{-5pt}
\paragraph{Discussion.} Our harmonious loss is related in spirit to uncertainty weighting \citep{kendall2018multitask} but has several key differences. Uncertainty weighting is derived from maximum likelihood estimation, which parameterizes noises of Gaussian-distributed outputs of each task, known as homoscedastic uncertainty. In contrast, our motivation is to balance loss scales among tasks. More specifically, measuring the uncertainty of observations and rewards results in putting each observation pixel on equal footing as the scalar reward, still overlooking the large disparity in dimension sizes. However, we take high-dimensional observations as a whole and directly balance the two losses. Furthermore, we do not make assumptions on the distributions behind losses, which makes it possible for us to balance the KL loss, while uncertainty weighting has no theoretical basis for doing so.

\vspace{-2pt}
\section{Experiments}
\label{sec:exp}

We evaluate the ability of HarmonyDream to boost sample efficiency of base MBRL methods on diverse and challenging visual control domains as shown in Figure~\ref{fig:example_obs_single}, including robotic manipulation and locomotion, and video game tasks.
We conduct most experiments for HarmonyDream based on DreamerV2 but also demonstrate its generality to other base MBRL methods, including DreamerV3 \citep{hafner2023mastering} and DreamerPro \citep{deng2022dreamerpro}.
Experimental details and additional results can be found in Appendix \ref{app:details} and \ref{app:additional experiments}.

\vspace{-1pt}
\subsection{Meta-world Experiments}
\label{sec:metaworld}

\vspace{-1pt}
\paragraph{Environment details.} 
Meta-world is a benchmark of 50 robotic manipulation tasks with fine-grained observation details, such as small target objects. 
Due to our limited computational resources, we choose a set of representative tasks according to the categories of task difficulty by \citet{seo2022masked}: two from the \textit{easy} category (Lever Pull and Handle Pull Side), two from the \textit{medium} category (Hammer and Sweep Into), and two from the \textit{hard} category (Push and Assembly). These tasks are run over different numbers of environment steps: \textit{easy} tasks and Hammer over 250K steps, Sweep Into over 500K steps, the else over 1M steps.

\begin{figure*}[!htbp]
    \centering
    \begin{subfigure}[t]{0.67\textwidth}
        \centering
        \includegraphics[width=\textwidth]{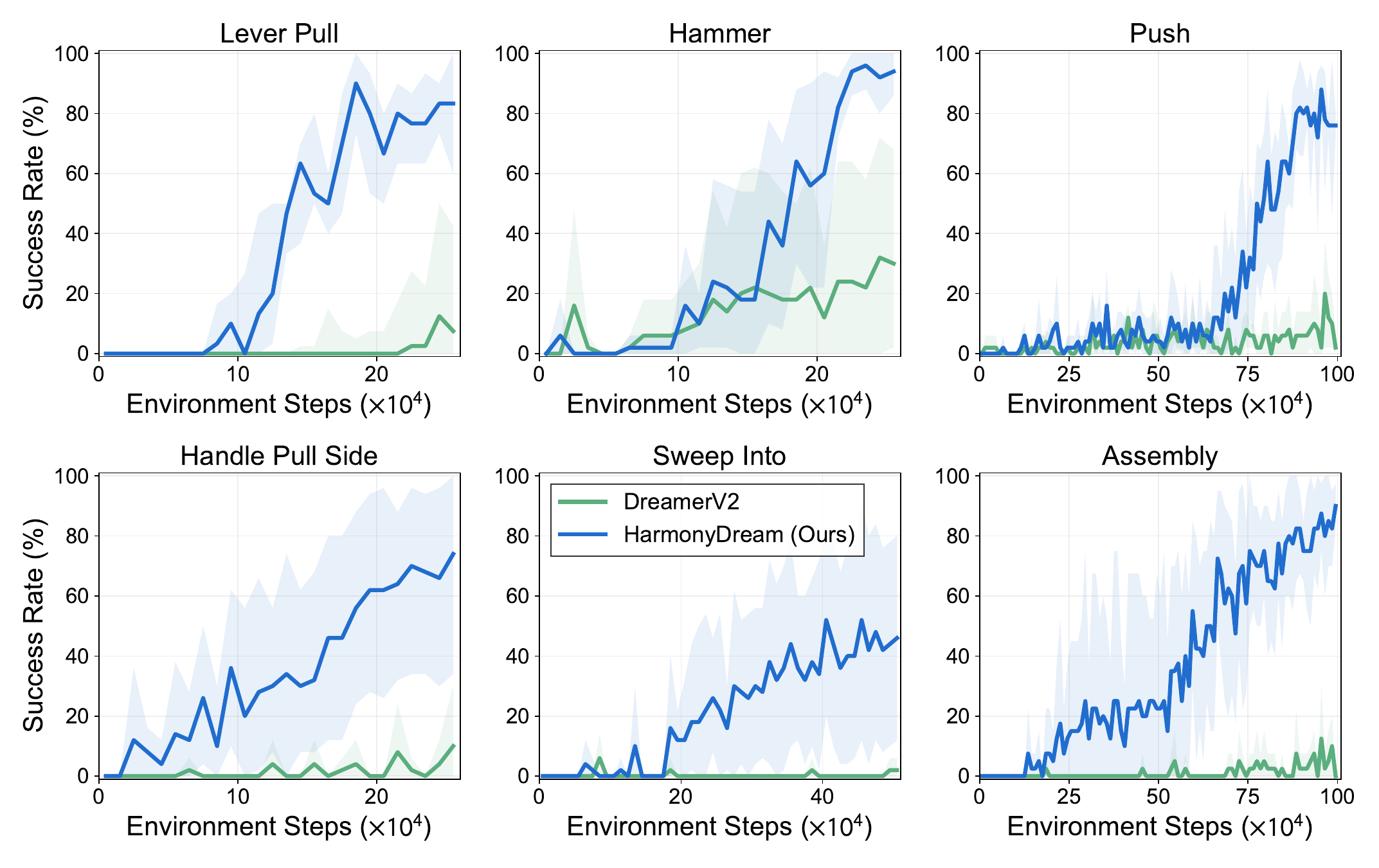}
        {
        \vspace{-17pt}
        \caption{Meta-world}
        \label{fig:metaworld_result}}
    \end{subfigure}
    ~
    \hspace{-10pt}
    \begin{subfigure}[t]{0.243\textwidth}
        \centering
        \includegraphics[width=\textwidth]{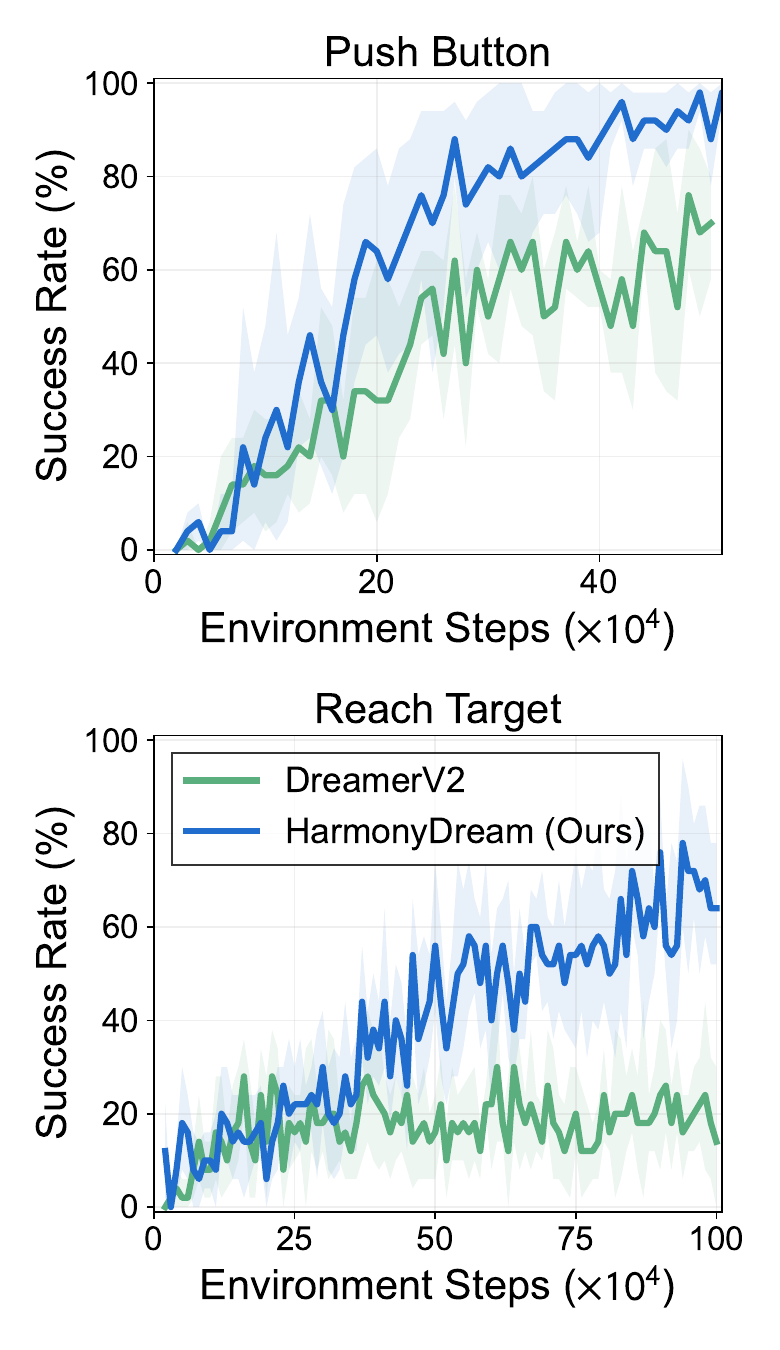}
        {
        \vspace{-17pt}
        \caption{RLBench}
        \label{fig:rlbench_result}}
    \end{subfigure}
    {
    \vspace{-5pt}
    \caption{Learning curves on visual manipulation tasks from (a) Meta-World and (b) RLBench benchmarks, measured on the success rate. We report the mean and 95\% confidence interval across five runs.
    \label{fig:manipulation_result}
    }}
    \vspace{-3pt}
\end{figure*}

\vspace{-9pt}
\paragraph{Results.} In Fig.~\ref{fig:metaworld_result}, we report the performance of HarmonyDream on six Meta-world tasks, in comparison with our base MBRL method DreamerV2. By simply adding harmonizers to the original DreamerV2 method, our HarmonyDream demonstrates superior performance in terms of both sample efficiency and final success rate. In particular, HarmonyDream achieves over 75\% and 90\% success rates on the challenging Push and Assembly tasks, respectively, while DreamerV2 fails to learn a meaningful policy.

\begin{figure*}[ht]
    \centering
    \begin{subfigure}[t]{0.69\textwidth}
        \centering
        \includegraphics[width=\textwidth]{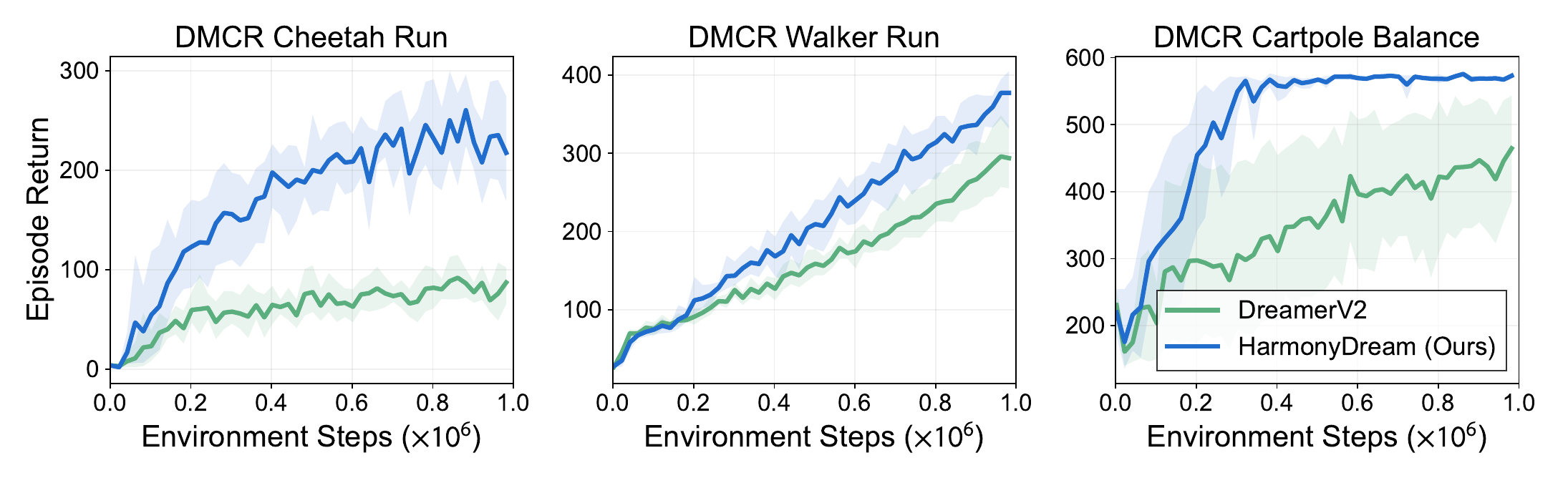}
        {
        \vspace{-17pt}
        \caption{Learning curves}
        \label{fig:dmcr_learning_curve}}
    \end{subfigure}
    ~
    \hspace{-10pt}
    \begin{subfigure}[t]{0.245\textwidth}
        \centering
        \includegraphics[width=\textwidth]{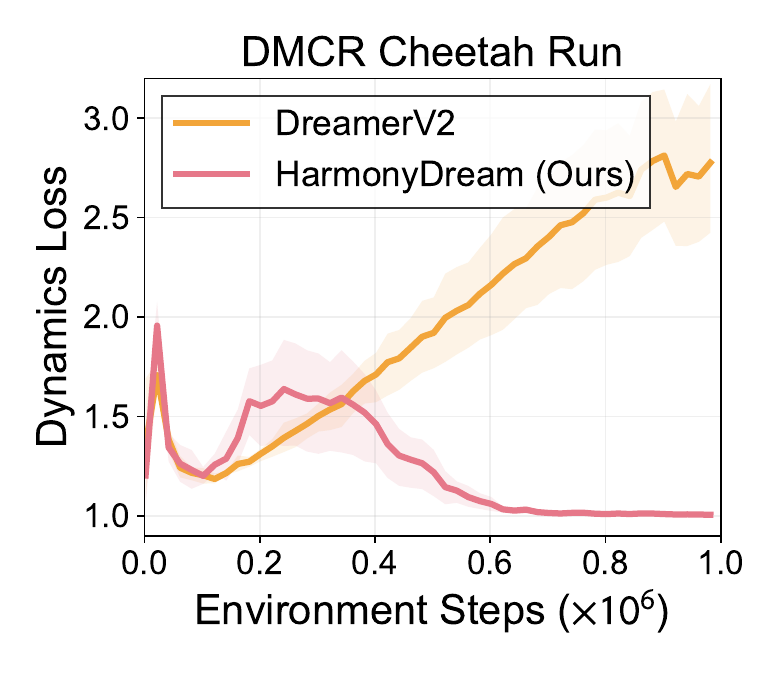}
        {
        \vspace{-17pt}
        \caption{Dynamics loss}
        \label{fig:kl_loss_cheetah_run}}
    \end{subfigure}
    {
    \vspace{-5pt}
    \caption{Learning curves (a) on three DMC Remastered visual locomotion tasks and (b) one dynamics loss curve shown on Cheetah Run. We report the mean and 95\% confidence interval across five runs. }
    \label{fig:dmcr_result}}
    \vspace{-5pt}
\end{figure*}

\subsection{RLBench Experiments}
\label{sec:rlbench}

\paragraph{Environment details.}
To assess our method on more complex visual robotic manipulation tasks, we perform evaluations on the RLBench \citep{james2020rlbench} domain. 
Most tasks in RLBench have high intrinsic difficulty and only offer sparse rewards. 
Learning these tasks requires expert demonstrations, dedicated network structure, and additional inputs \citep{james2022qattention, james2022coarse}, which is out of our scope. 
Therefore, following \citet{seo2022masked}, we conduct experiments on two relatively easy tasks (Push Button and Reach Target) with dense rewards.

\vspace{-8pt}
\paragraph{Results.}
In Fig.~\ref{fig:rlbench_result}, we show the superiority of our approach on the RLBench domain. HarmonyDream offers 28\% of absolute final performance gain on the Push Button task and 50\% on the more difficult Reach Target tasks. The results presented above prove the ability of HarmonyDream to promote sample efficiency of model-based RL on robotic manipulation domains for both easy and difficult tasks.

\subsection{DMC Remastered Experiments}
\label{sec:dmcr}

\paragraph{Environment details.}
DMC Remastered \cite{grigsby2020measuring} is a challenging extension of the widely used robotic locomotion benchmark, DeepMind Control Suite \citep{tassa2018deepmind} with randomly generated graphics emphasizing visual diversity. We train and evaluate our agents on three tasks: Cheetah Run, Walker Run, and Cartpole Balance.

\vspace{-8pt}
\paragraph{Results.}
Fig.~\ref{fig:dmcr_learning_curve} demonstrates the effectiveness of HarmonyDream on three DMCR tasks. Our method greatly enhances the base DreamerV2 method to unleash its potential. Fig.~\ref{fig:kl_loss_cheetah_run} shows different learning curves of the dynamics loss between HarmonyDream and DreamerV2. It is worth noting that DMCR tasks contain distracting visual factors, such as background and robot body color, which may hinder the learning process of observation modeling. DreamerV2 diverges in learning loss on this task, but by leveraging the importance of reward modeling, HarmonyDream bypasses distractors in observations and can learn task-centric transitions more easily, indicated by converged dynamics loss.

\begin{figure*}[htbp]
    \centering
    \begin{subfigure}[t]{0.47\textwidth}
        \centering
        \includegraphics[width=\textwidth]{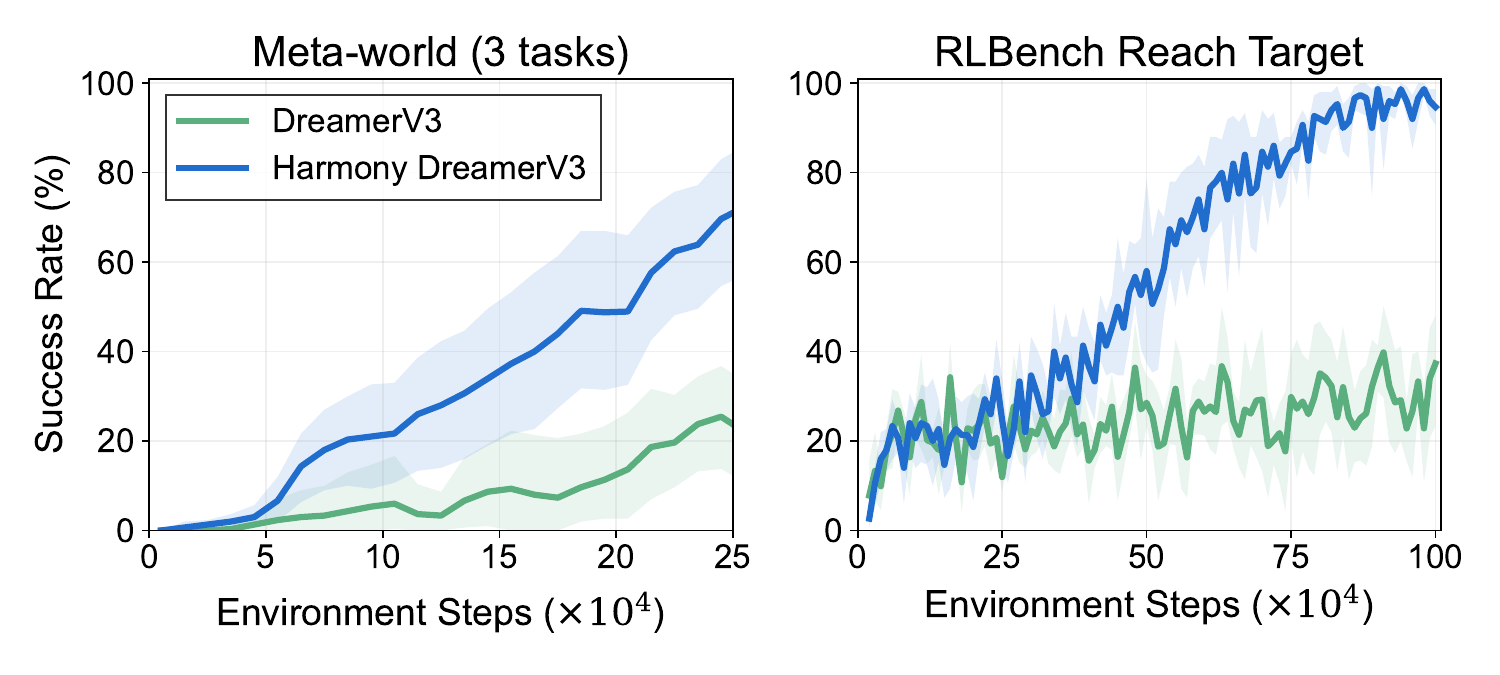}
    \end{subfigure}
    ~
    \hspace{-10pt}
    \begin{subfigure}[t]{0.47\textwidth}
        \centering
        \includegraphics[width=\textwidth]{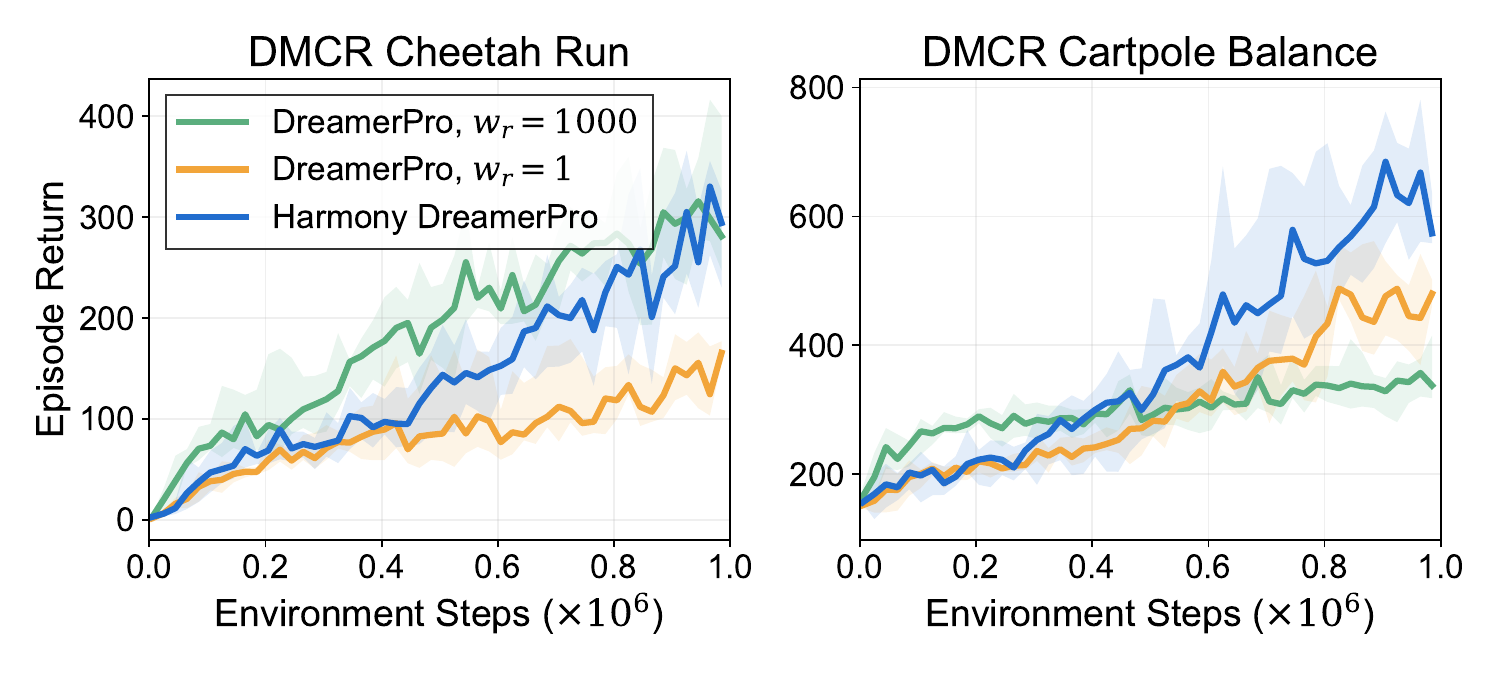}
    \end{subfigure}
    {
    \vspace{-5pt}
    \caption{Performance of HarmonyDream applied to DreamerV3 (\textit{left}) and DreamerPro (\textit{right}).}
    \label{fig:model_generality}}
    \vspace{-2pt}
\end{figure*}

\subsection{Generality to Model-based RL Methods}
\label{sec:generality}

\begin{figure}[t]
    \centering
    \begin{subfigure}[t]{0.25\textwidth}
        \centering
        \raisebox{0.07\height}{
        \includegraphics[width=\textwidth]{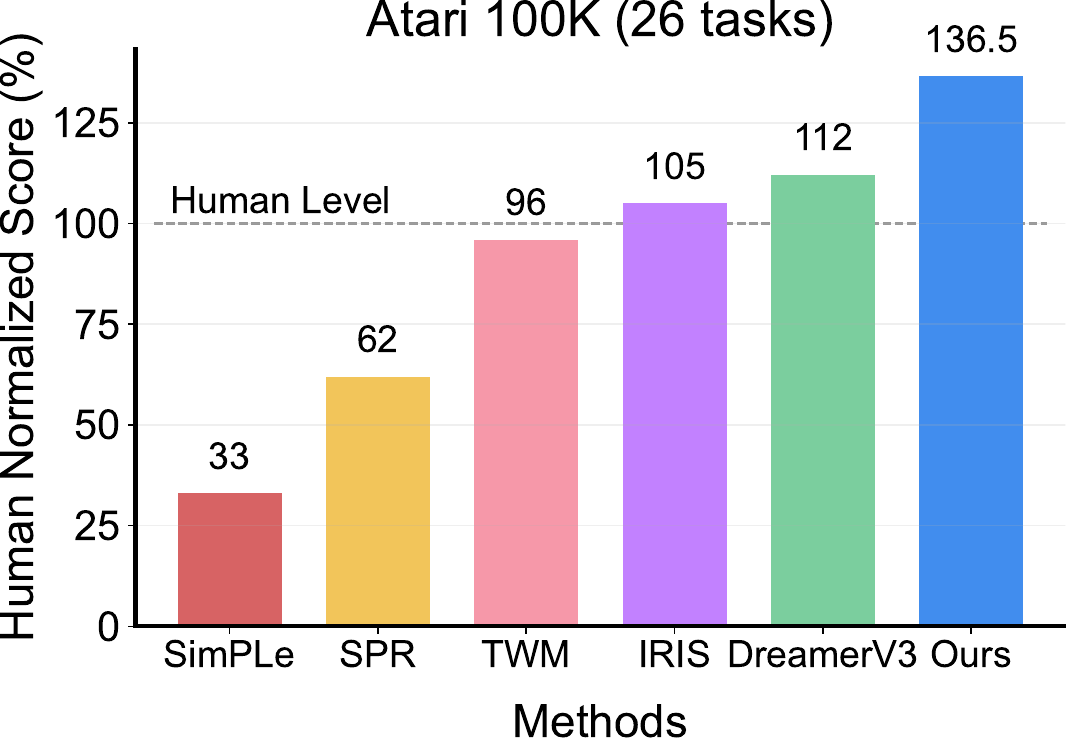}}
    \end{subfigure}
    ~
    \hspace{-10pt}
    \begin{subfigure}[t]{0.23\textwidth}
        \centering
        \includegraphics[width=\textwidth]{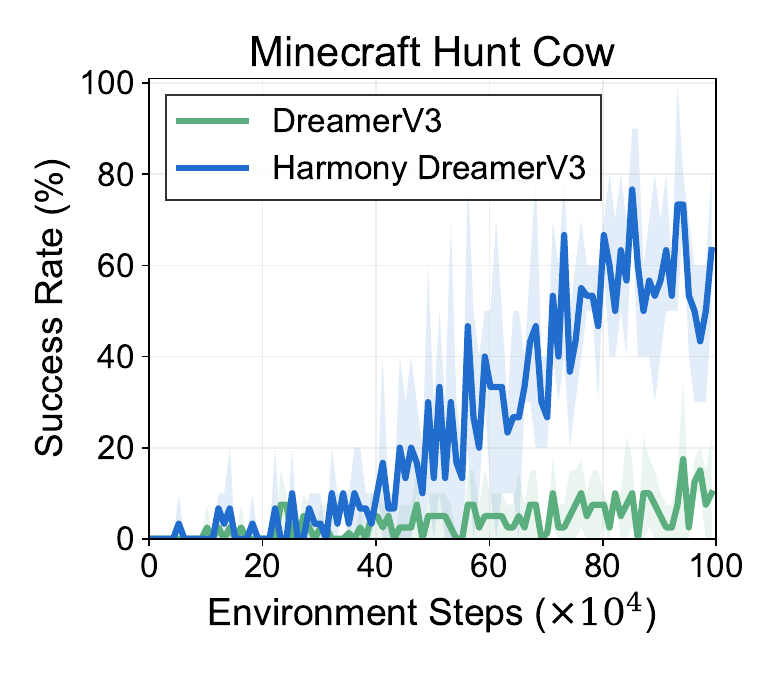}
    \end{subfigure}
    {
    \vspace{-20pt}
    \caption{Performance of HarmonyDream based on DreamerV3 on the Atari 100K benchmark (\textit{left}) and the Hunt Cow task from the MineDojo benchmark (\textit{right}).}
    \label{fig:atari_minecraft}}
    \vspace{-10pt}
\end{figure}

\paragraph{DreamerV3.} DreamerV3 \citep{hafner2023mastering} improves DreamerV2 to master diverse domains. Notably, our method is orthogonal to the various modifications in DreamerV3. 
DreamerV3 introduces a static symlog transformation to mitigate the problem of different per-dimension scales across environment domains, while HarmonyDream dynamically balances the overall loss scales across tasks in world model learning, considering together per-dimension scales, dimensions, and training dynamics. We refer to a detailed discussion in Appendix \ref{app:difference_dreamerv3}. Experiments on Meta-world and RLBench, as shown in Fig.~\ref{fig:model_generality}, illustrate that our method can combine with DreamerV3 to further improve performance. To further illustrate the applicability of our method, we also evaluate our Harmony DreamerV3 on two video game domains: Minecraft and Atari. For the Minecraft domain, we choose a challenging task of learning a basic skill, Hunt Cow, from the MineDojo benchmark \citep{fan2022minedojo}. As shown in Fig.~\ref{fig:atari_minecraft}, Harmony DreamerV3 exhibits great improvement in the Minecraft domain. For the Atari 100K benchmark \cite{kaiser2019model}, we improve DreamerV3 to achieve 136.5\% of mean human performance, setting a new state of the art among methods without lookahead search.

\begin{figure*}[htbp]
    \centering
    \begin{subfigure}[t]{0.24\textwidth}
        \centering
        \includegraphics[width=\textwidth]{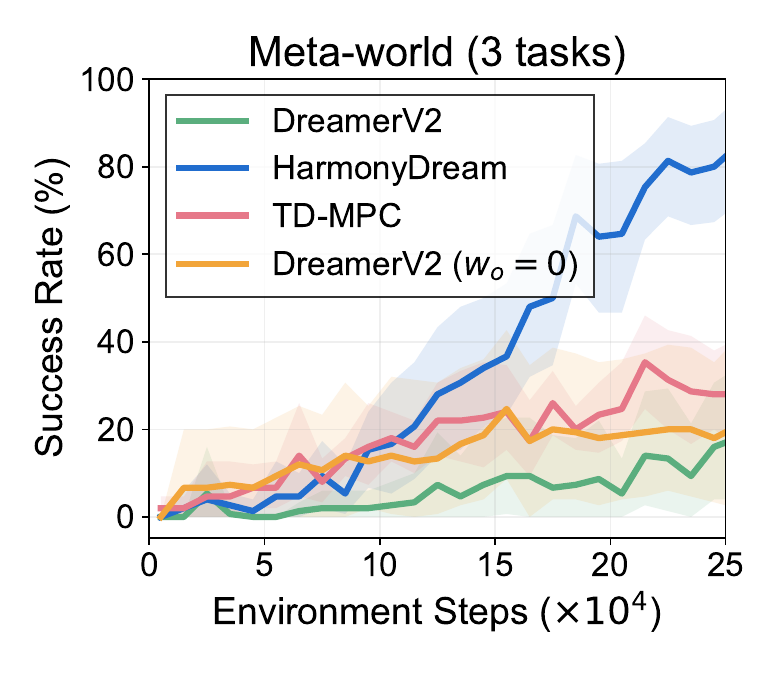}
        {
        \vspace{-15pt}
        \caption{Comparison to TD-MPC}
        \label{fig:tdmpc_dmcr_aggregated}}
    \end{subfigure}
    \hspace{-5pt}
    \begin{subfigure}[t]{0.72\textwidth}
        \centering
        \includegraphics[width=0.333\textwidth]{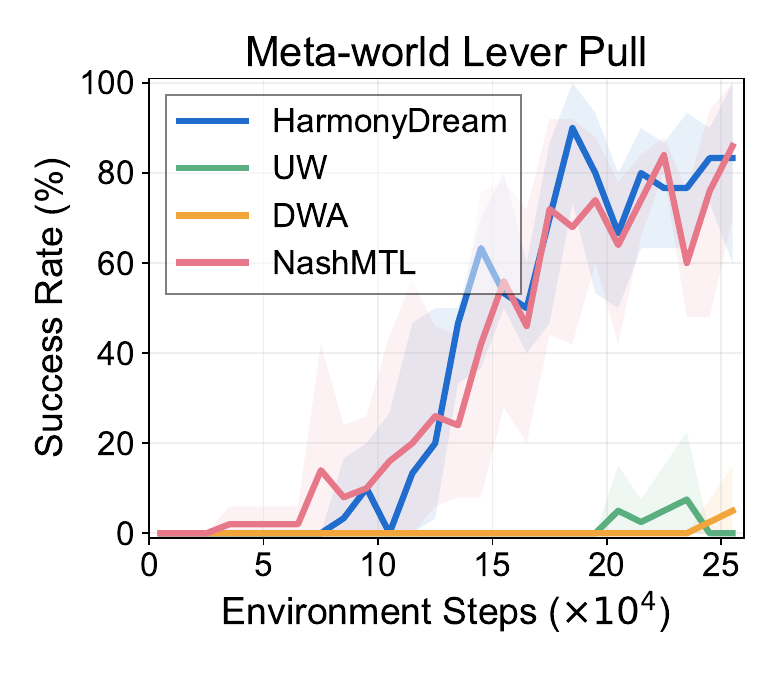}
        \hspace{-8pt}
        \includegraphics[width=0.333\textwidth]{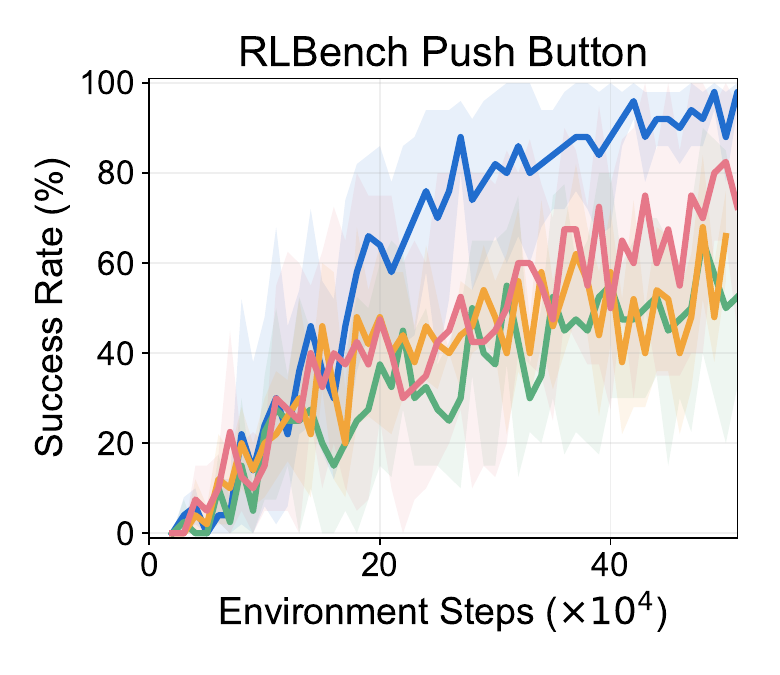}
        \hspace{-8pt}
        \includegraphics[width=0.333\textwidth]{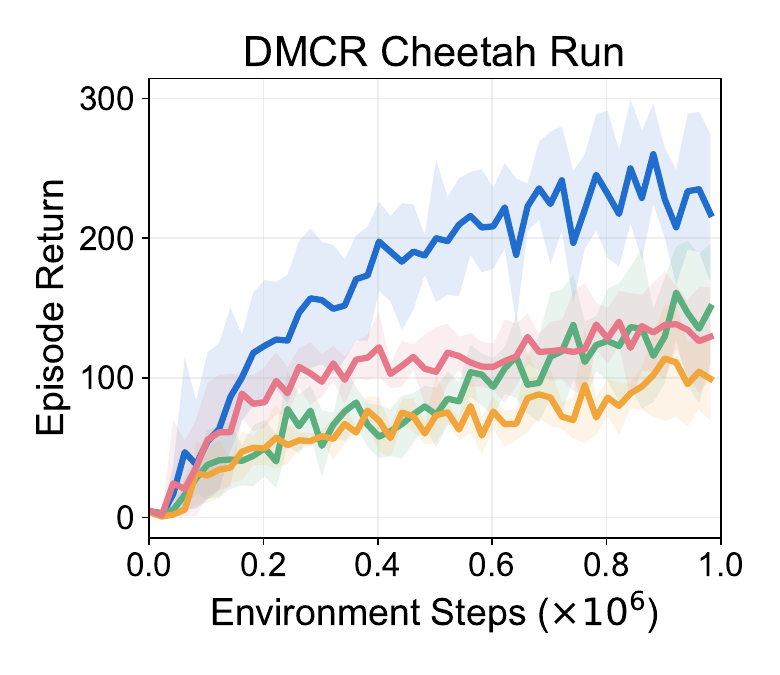}
        {
        \vspace{-3pt}
        \caption{Comparison to multi-task learning methods applied to DreamerV2}
        \label{fig:multi-task-baseline}}
    \end{subfigure}
    {
    \caption{Comparison of HarmonyDream to implicit MBRL methods and multi-task learning methods.}
    }
    \vspace{8pt}
\end{figure*}

\vspace{-8pt}
\paragraph{DreamerPro.} DreamerPro \citep{deng2022dreamerpro} is a model-based RL method that ``reconstructs" the cluster assignment of the observation. We conduct DreamerPro experiments on the DMCR domain. By default, DreamerPro uses a manually tuned reward loss weight $w_r=1000$.
We demonstrate in Fig.~\ref{fig:model_generality} that our method can still achieve higher sample efficiency and, on average, outperform manually tuned loss weights that are computationally costly. 

\subsection{Analysis}

\paragraph{Comparison to implicit MBRL.}
As shown in Sec.~\ref{sec:observaions}, learning from reward modeling alone lacks sample efficiency. However, one may argue that purposefully designed implicit MBRL methods can be more effective. In Fig.~\ref{fig:tdmpc_dmcr_aggregated}, we show comparisons with an implicit MBRL method, TD-MPC \citep{hansen2022temporal} on three tasks of Meta-world. We observe that TD-MPC has difficulty in efficient learning as it lacks observation modeling to guide representation learning. In contrast, our method achieves superior performance. We also compare with another implicit MBRL method, RePo \cite{zhu2023repo}, as shown in the following paragraph.

\vspace{-8pt}
\paragraph{Comparison to Dreamer-based task-centric methods.}

\begin{figure*}[ht]
    \vspace{-5pt}
    \centering
    \begin{subfigure}[t]{0.47\textwidth}
        \centering
        \includegraphics[width=\textwidth]{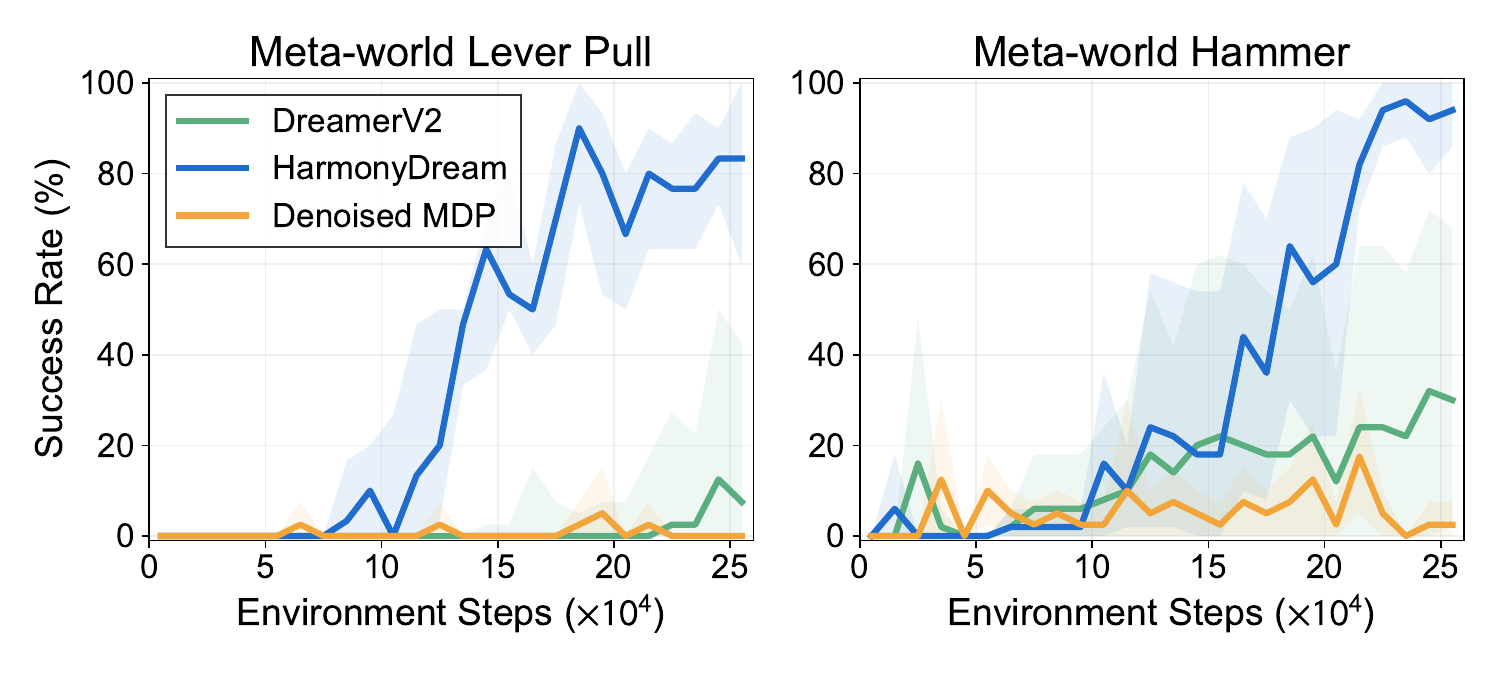}
    \end{subfigure}
    ~
    \hspace{-10pt}
    \begin{subfigure}[t]{0.243\textwidth}
        \centering
        \includegraphics[width=\textwidth]{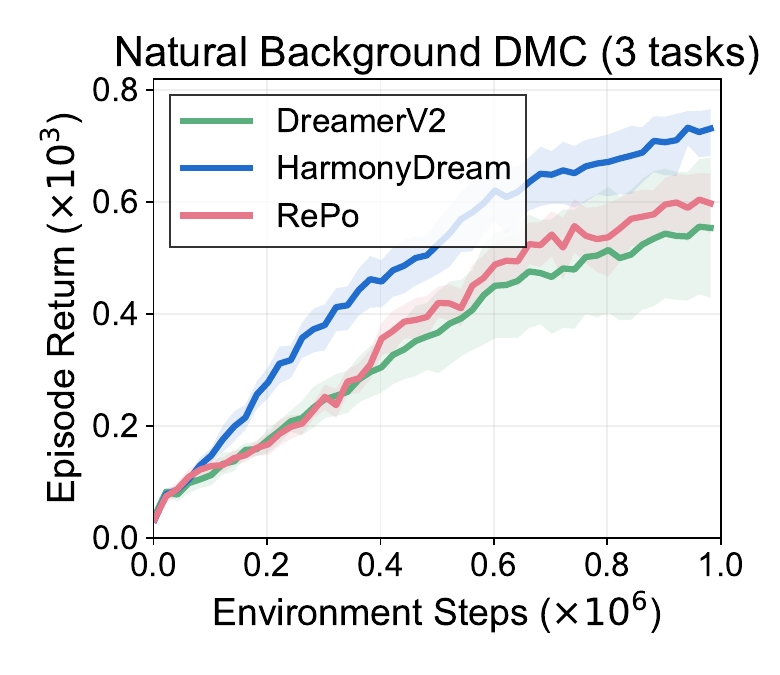}
    \end{subfigure}
    \begin{subfigure}[t]{0.243\textwidth}
        \centering
        \includegraphics[width=\textwidth]{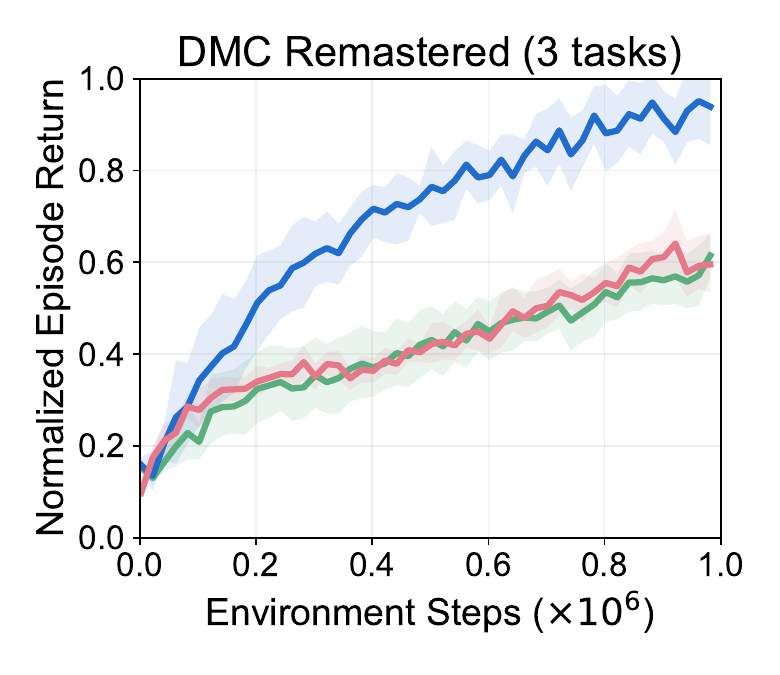}
    \end{subfigure}
    {\vspace{-10pt}
    \caption{Comparison of HarmonyDream with Dreamer-based task-centric methods, Denoised MDP (\textit{left}) and RePo (\textit{right}).}
    \vspace{-10pt}
    \label{fig:comparison_dreamer_based_results}}
\end{figure*}

Denoised MDP \citep{wang2022denoisedmdps} and RePo \citep{zhu2023repo} represent modifications to the Dreamer architecture that share a similar point with our approach in enhancing task-centric representations. We compare our method to these two methods on Meta-world, DMC Remastered, and additionally, natural background DMC \citep{zhang2018natural}, which is also a distracted DMC variant used originally in the RePo paper. Fig.~\ref{fig:comparison_dreamer_based_results} shows that our HarmonyDream has a higher sample efficiency than Denoised MDP and RePo. Detailed discussion and comparison to these methods can be found in Appendix \ref{app:comparison_denoised_mdp} and \ref{app:comparison_repo}, respectively.

\vspace{-8pt}
\paragraph{Comparison to multi-task learning methods.} While our focus is not on developing a new multi-task learning method, we compare HarmonyDream with advanced methods in this area, including Uncertainty Weighting (UW, \citet{kendall2018multitask}), Dynamics Weight Average (DWA, \citet{liu2019endtoend}), and NashMTL \cite{navon2022multitask}. Fig.~\ref{fig:multi-task-baseline} illustrates that our straightforward method is the most effective one among these methods, which also has the advantage of extreme simplicity. In-depth discussions on the differences between methods and why these methods can hardly make more improvements are included in Appendix \ref{app:multi-task-baseline}.

\vspace{-8pt}
\paragraph{Ablation on rectified loss.} We illustrate, through Fig.~\ref{fig:unrectified_HarmonyDream_dmc} in Appendix, the effectiveness of our rectified loss (Eq.~(\ref{eq:harmony_loss_with_base})) in enhancing training stability and final performance.

\section{Related Work}

\paragraph{World models for visual RL.}
There exist several approaches to learning world models that explicitly model observations, transitions, and rewards. They can be widely utilized to boost sample efficiency in visual RL. In world models, visual representation can be learned via image reconstruction \citep{ha2018recurrent, kaiser2019model, hafner2019learning, seo2022masked, seo2022reinforcement}, or reconstruction-free contrastive learning \citep{okada2021dreaming,deng2022dreamerpro}.
Dreamer \citep{hafner2019dream, hafner2020mastering, hafner2023mastering} represents a series of methods that learn latent dynamics models from observations and learn behaviors by latent imagination. These methods have proven their effectiveness in tasks like video games \citep{hafner2020mastering} and real robot control \citep{wu2022daydreamer}.
Regardless, the problem of task domination is general for world models, and our findings and approach are not limited to our focused Dreamer architecture.

\vspace{-6pt}
\paragraph{Implicit model-based RL.}
Implicit MBRL \citep{moerland2023survey} is a more abstract approach and aims to learn value equivalence models \citep{grimm2020value} that focus on task-centric characteristics of the environment. This approach mitigates the objective mismatch \citep{lambert2020objective} between maximum likelihood estimation for world models and maximizing returns for policies. A typical success is MuZero \citep{schrittwieser2020mastering, ye2021mastering}, which learns a world model by predicting task-specific rewards, values, and policies, without observation reconstruction. Similarly, TD-MPC \cite{hansen2022temporal} learns implicit world models for continuous control. While focusing on Dreamer, our analysis is consistent with those of MuZero showing that the potential efficiency of task-centric models can be better released when properly leveraging richer information from observation models \citep{anand2021procedural}.

\vspace{-6pt}
\paragraph{Multi-task learning.}
Multi-task learning \citep{caruana1997multitask, ruder2017overview} aims to improve different tasks by jointly learning from a shared representation. A common approach is to aggregate task losses, where the loss or gradient of each task is manipulated by criteria like uncertainty \citep{kendall2018multitask}, performance metric \citep{guo2018dynamic}, gradient norm \citep{chen2018gradnorm} or gradient direction \citep{yu2020gradient, wang2020gradient, navon2022multitask}, to avoid negative transfer \citep{jiang2023forkmerge}. 
Previous works on multi-task learning in RL typically considered different policy learning tasks defined by different reward functions or environment dynamics \citep{rusu2016progressive, teh2017distral, yu2020gradient}. In contrast, we innovatively depict world model learning as multi-task learning, composed of reward and observation modeling, and HarmonyDream learns to maintain a delicate equilibrium between them to mitigate domination.

\section{Discussion}
\label{sec:discussion}

We identify two tasks inside world models---observation and reward modeling---and interpret different MBRL methods as different task weighting. Our empirical study reveals that domination of a particular task can dramatically deteriorate the sample efficiency of MBRL. We thus introduce HarmonyDream, a simple world model learning approach that dynamically balances these tasks, thereby substantially improving sample efficiency.

HarmonyDream is particularly effective for scenarios where observation models are necessary for better representation learning, but the default weighting strategy of explicit world model learning causes negative effects due to observation modeling domination. These scenarios are mainly vision-based RL tasks, typically with complicated observations, including but not limited to:
\begin{itemize}
    \item \textbf{Fine-grained task-relevant observations}: Robotics manipulation tasks (e.g., Meta-world and RLBench) and video games (e.g., Atari games, particularly Breakout, Qbert, and Gopher) require accurately modeling interactions with small objects.
    \item \textbf{Highly varied task-irrelevant observations}: Redundant visual components such as backgrounds (e.g., natural background DMC) and body color (e.g., DMCR) can easily distract visual agents if task-relevant information is not emphasized correctly.
    \item \textbf{Hybrid of both}: More difficult open-world tasks (e.g. Minecraft) can encounter both, including small target entities and abundant visual details.
\end{itemize}

These environment features are ubiquitous in realistic applications, and simply emphasizing reward modeling through HarmonyDream without any architecture modifications or hyperparameter tuning can make remarkable improvements.

Benchmark environments featuring clean observations with prominent target objects, such as standard DMC and Crafter \citep{hafner2022benchmarking}, do not encounter significant domination of observation modeling and are expected to gain marginal improvements with HarmonyDream, as shown in Fig.~\ref{fig:dmc_result} (for DMC) and Fig.~\ref{fig:dreamerv3_crafter} (for Crafter) in Appendix. Nevertheless, we do not observe any negative performance change with HarmonyDream on these clean benchmarks.

The development of our method is primarily based on empirical and intuitive observations. A future direction is to explain our method theoretically, or to better measure and balance the contributions of world model tasks empirically, beyond simply considering loss scales.
We hope our work can offer valuable insights and help pave the way for exploiting the multi-task nature of world models.

\section*{Acknowledgements}

We would like to thank many colleagues, in particular Haixu Wu, Baixu Chen, Yuhong Yang, Chaoyi Deng, and Jincheng Zhong, who provided us with valuable discussions. This work was supported by the National Natural Science Foundation of China (62022050 and U2342217), the BNRist Innovation Fund (BNR2024RC01010), the Huawei Innovation Fund, and the National Engineering Research Center for Big Data Software.

\section*{Impact Statement}

This paper presents work whose goal is to advance the field of 
Machine Learning. There are many potential societal consequences 
of our work, none which we feel must be specifically highlighted here.


\bibliography{icml2024_conference}
\bibliographystyle{icml2024}

\newpage
\appendix
\onecolumn
\section{Behavior Learning}
\label{app:behavior learning}

HarmonyDream does not change the behavior learning procedure of its base MBRL methods \citep{hafner2020mastering, hafner2023mastering, deng2022dreamerpro}, and we briefly describe the actor-critic learning scheme shared with these base methods. 

Specifically, we leverage a stochastic actor and a deterministic critic parameterized by $\psi$ and $\xi$, respectively, as shown below:
\begin{equation}
    \label{eq:behavior learning}
    \text { Actor: } \hat{a}_t \sim \pi_\psi\left(\hat{a}_t \mid \hat{z}_t\right) \quad \text { Critic: } v_{\xi}\left(\hat{z}_t\right) \approx \mathrm{E}_{p_\theta, \pi_\psi}\left[\sum\nolimits_{\tau \geq t} \gamma^{\tau-t} \hat{r}_\tau\right],
\end{equation}
where $p_\theta$ is the world model. The actor and critic are jointly trained on the same imagined trajectories $\{ \hat{z}_\tau, \hat{a}_\tau, \hat{r}_\tau\}$ with horizon $H$, generated by the transition model and reward model in Eq.~(\ref{eq:dreamer}) and the actor in Eq.~(\ref{eq:behavior learning}). The critic is trained to regress the $\lambda$-target:
\begin{align}
    &\mathcal{L}_{\text{critic}}(\xi)\doteq\mathbb{E}_{p_{\theta}, \pi_\psi}\left[\sum^{t+H}_{\tau=t} \frac{1}{2} \left(v_{\xi}(\hat{z}_{\tau}) - \text{sg}(V_{\tau}^{\lambda})\right)^{2}\right],
    \label{eq:critic_loss}\\
    &V_{\tau}^{\lambda}\doteq \hat{r}_{\tau} + \gamma
    \begin{cases}
      (1 - \lambda)v_{\xi}(\hat{z}_{\tau+1})+\lambda V_{\tau+1}^{\lambda} & \text{if}\ \tau<t+H \\
      v_{\xi}(\hat{z}_{\tau + 1}) & \text{if}\ \tau=t+H.
    \end{cases}
\end{align}
The actor, meanwhile, is trained to output actions that maximize the critic output by backpropagating value gradients through the learned world model. The actor loss is defined as follows:
\begin{align}
    \mathcal{L}_{\text{actor}}(\psi)\doteq \mathbb{E}_{p_{\theta}, \pi_\psi} \left[\sum^{t+H}_{\tau=t} \left(-V_{\tau}^{\lambda} - \eta\,\text{H}\left[\pi_\psi (\hat{a}_{\tau}|\hat{z}_{\tau}) \right] \right) \right],
\end{align}
where $\text{H}\left[\pi_\psi (\hat{a}_{\tau}|\hat{z}_{\tau}) \right]$ is an entropy regularization which encourages exploration, and $\eta$ is the hyperparameter that adjusts the regularization strength. For more details, we refer to \citet{hafner2019dream}.

\section{Derivations}
\label{app:derivations}

\paragraph{Proof of Proposition \ref{prop:harmony}.} To minimize $\mathbb{E}[\mathcal{H}(\mathcal{L}, \sigma)]$, we force the the partial derivative w.r.t. $\sigma$ to $0$:
\begin{align}
    \nabla_{\sigma}\mathbb{E}[\mathcal{H}(\mathcal{L}, \sigma)] &= \nabla_{\sigma}\mathbb{E}\left[\frac{1}{\sigma}\mathcal{L} + \log{\sigma}\right] = \mathbb{E}\left[ \nabla_\sigma \left(\frac{1}{\sigma}\mathcal{L} + \log{\sigma}\right)\right] \\
    &= \mathbb{E}\left[ -\frac{1}{\sigma^2}\mathcal{L} + \frac{1}{\sigma}\right] = \frac{1}{\sigma} -\frac{1}{\sigma^2}\mathbb{E}[\mathcal{L}] = 0.
\end{align}
This results in the solution $\sigma^*=\mathbb{E}[\mathcal{L}]$, and equivalently, the harmonized loss scale is $\mathbb{E}[\mathcal{L}/\sigma^*]=1$.

\paragraph{Analytic solution of rectified loss.} Similarly, minimizing $\mathbb{E}\big[\hat{\mathcal{H}}(\mathcal{L}, \sigma)\big]$ yields
\begin{gather}
\begin{aligned}
    \nabla_\sigma \mathbb{E}\left[\hat{\mathcal{H}}(\mathcal{L}, \sigma) \right]
        &= \nabla_\sigma\left(\frac{1}{{\sigma}}\mathbb{E}[\mathcal{L}] +\log{\left(1+\sigma\right)}\right) = -\frac{1}{{\sigma}^2}\mathbb{E}[\mathcal{L}] + \frac{1}{1+\sigma} =0 \\
     \sigma&=\frac{\mathbb{E}[\mathcal{L}]+\sqrt{{\mathbb{E}[\mathcal{L}]}^2+4\mathbb{E}[\mathcal{L}]}}{2}.&
\end{aligned}
\end{gather}
Therefore the learnable loss weight, in our rectified harmonious loss, approximates the analytic loss weight: 
\begin{equation}
    \frac{1}{\sigma} = \frac{2}{\mathbb{E}[\mathcal{L}]+\sqrt{{\mathbb{E}[\mathcal{L}]}^2+4\mathbb{E}[\mathcal{L}]}},
\end{equation}
corresponding to a loss scale $\mathbb{E}[\mathcal{L}]$, which is less than the unrectified $1/\mathbb{E}[\mathcal{L}]$. Adding a constant in the regularization term $\log(1+\sigma)$ results in the $4\mathbb{E}[\mathcal{L}]$ in the $\sqrt{{\mathbb{E}[\mathcal{L}]}^2+4\mathbb{E}[\mathcal{L}]}$ term, which prevents the loss weight from getting extremely large when faced with a small $\mathbb{E}[\mathcal{L}]$.

\section{Experimental Details}
\label{app:details}

\subsection{Benchmark Environments}

\paragraph{Meta-world.}
Meta-world \citep{yu2020meta} is a benchmark of 50 distinct robotic manipulation tasks. We choose six tasks in all according to the difficulty criterion \textit{(easy, medium, hard, and very hard)} proposed by \citet{seo2022masked}. Specifically, we choose Handle Pull Side and Lever Pull from the \textit{easy} category, Hammer and Sweep Into from the \textit{medium} category, and Push and Assembly from the \textit{hard} category. We observe that although the Hammer task belongs to the \textit{medium} category, it is relatively easy for the DreamerV2 agent to learn, and our HarmonyDream can already achieve high success with 250K environment steps. Therefore, we train our agents over 250K environment steps on Hammer, along with the two \textit{easy} tasks. For the remaining tasks, we train our agents over 500K environment steps for Sweep Into, and 1M environment steps for Push and Assembly, according to their various difficulties. In all tasks, the episode length is 500 environment steps with no action repeat.

\begin{figure*}[htbp]
    \centering
    \includegraphics[width=0.8\textwidth]{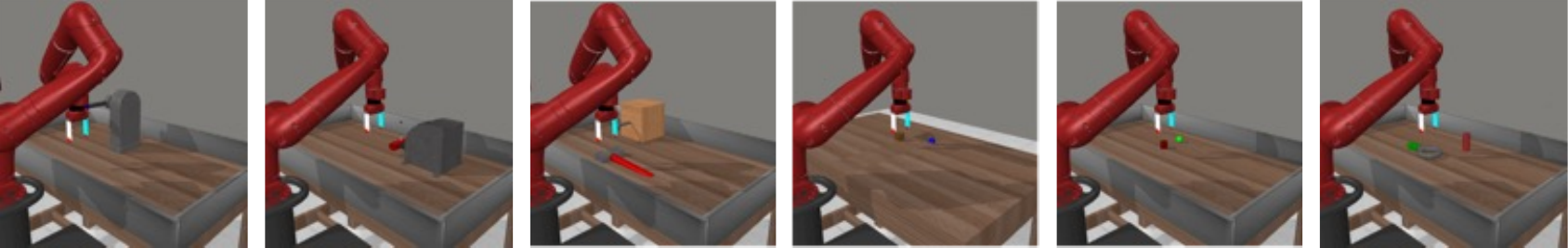}
    \caption{Example observations of Meta-world tasks: Lever Pull, Handle Pull Side, Hammer, Sweep Into, Push, and Assembly (left to right).}
    \label{fig:example_obs_metaworld}
\end{figure*}

\begin{wrapfigure}{r}{0.36\textwidth}
\vspace{-8pt}
\begin{center}
\centerline{\includegraphics[width=0.26\textwidth]{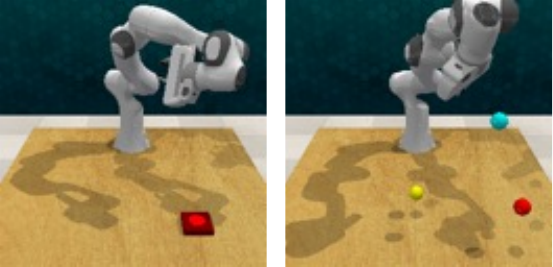}}
\caption{Example observations of RLBench tasks: Push Button and Reach Target.}
\label{fig:example_obs_rlbench}
\vspace{-17pt}
\end{center}
\end{wrapfigure}

\paragraph{RLBench.}
RLBench \citep{james2020rlbench} is a challenging benchmark for robot learning.  Most tasks in RLBench are overchallenging for DreamerV2, even equipped with HarmonyDream. Therefore, following \citet{seo2022masked}, we choose two relatively easy tasks (i.e. Push Button, Reach Target) and use an action mode that specifies the delta of joint positions.
Because the original RLBench benchmark does not provide dense rewards for the Push Button task, we assign a dense reward following \citet{seo2022masked}, which is defined as the sum of the L2 distance of the gripper to the button and the magnitude of the button being pushed. 
In our experiments, we found that the original convolutional encoder and decoder of DreamerV2 can be insufficient for learning the RLBench task. Therefore, in this domain, we adopt the ResNet-style encoder and decoder from \citet{wu2023pretraining} for both DreamerV2 and our HarmonyDream. Note here that changes in the encoder and decoder architecture are completely orthogonal to our method and contributions. For tasks in the RLBench domain, the maximum episode length is set to 400 environment steps with an action repeat of 2.

\begin{wrapfigure}{r}{0.36\textwidth}
\vspace{-10pt}
\begin{center}
\centerline{\includegraphics[width=0.26\textwidth]{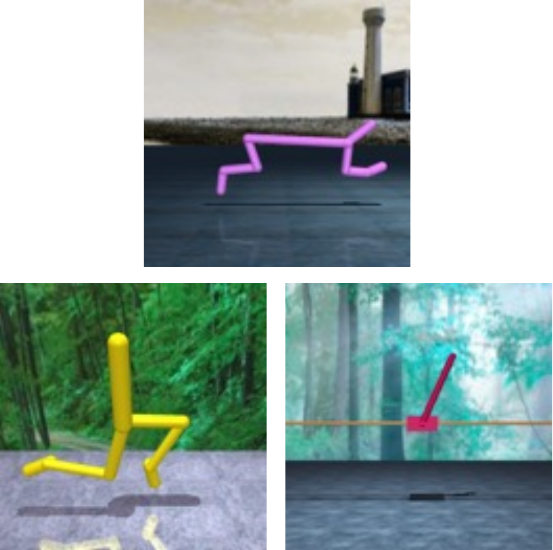}}
\caption{Example observations of DMC Remastered tasks: Cheetah Run, Walker Run, and Cartpole Balance.}
\label{fig:example_obs_dmcr}
\vspace{-30pt}
\end{center}
\end{wrapfigure}

\paragraph{DMC Remastered.}
The DMC Remastered (DMCR) \citep{grigsby2020measuring} benchmark is a challenging extension of the widely used robotic locomotion benchmark, DeepMind Control Suite \citep{tassa2018deepmind}, by expanding a complicated graphical variety. On initialization of each episode for both training and evaluation, the DMCR environment randomly resets 7 factors affecting visual conditions, including floor texture, background, robot body color, target color, reflectance, camera position, and lighting. Our agents are trained and evaluated on three tasks: Cheetah Run, Walker Run, and Cartpole Balance. We use all variation factors in all of our experiments and train our agents over 1M environment steps. Following the common setup of DeepMind Control Suite \citep{hafner2019dream, yarats2021mastering}, we set the episode length to 1000 environment steps with an action repeat of 2.

\paragraph{Atari 100K Benchmark.} The Atari 100K benchmark contains 26 video games with up to 18 discrete actions. On this benchmark, the agent is allowed to interact with each game environment for 100K steps, equivalent to 400K frames due to a frameskip of 4. This number of interaction steps, roughly two hours of real-time gameplay, has become a widely adopted standard in the realm of sample-efficient reinforcement learning. Human players are evaluated after two hours to get familiar with the game. Following the established protocol, we report the raw performance for each game, and the mean and median of human normalized scores: $\left(\mathrm{score}_{\mathrm{agent}}-\mathrm{score}_{\mathrm{random }}\right) /\left(\mathrm{score}_{\mathrm{human}}-\mathrm{score}_{\mathrm{random }}\right)$. For this benchmark, we keep all implementation details the same as DreamerV3.

\paragraph{Natural Background DMC.}
Natural background DMC \citep{zhang2018natural} modifies the DeepMind Control Suite by substituting its static background with natural videos. In our paper, this environment is implemented using the RePo \citep{zhu2023repo} codebase\footnote{\url{https://github.com/zchuning/repo}}. Following RePo, we train and evaluate our agent on three tasks: Cheetah Run, Walker Run and Cartpole Swingup. We adopt the standard configuration of DMC for natural background DMC, with a maximum episode length of 1000 environment steps and an action repeat of 2. 

\paragraph{Minecraft.}
Minecraft is a popular open-world game where a player explores a procedurally generated 3D world with diverse types of terrains to roam, materials to mine, tools to craft, structures to build, and wonders to discover. We leverage MineDojo \citep{fan2022minedojo}, an massive simulation suite developed on Minecraft, encompassing over 3000 distinct tasks. Our focus was to master a fundamental skill, \textit{Hunt Cow}, utilizing the manual dense reward provided by MineDojo. We prune the action space of MineDojo to Table \ref{tab:actspace_minedojo}, following the practice of STG-Transformer \citep{zhou2023learning}. For this benchmark, we employ the \textit{Large} model size variant of DreamerV3, comprising approximately 77M parameters. To ensure the terrain diversity of the environment, we hard reset the environment to generate a new world every 5 episodes. Observations for our agents consist solely of RGB frames, with a resolution of $128\times128\times3$ pixels. The maximum episode length is 500 environment steps, with no action repeat.

\begin{table}[h]
\centering
\caption{Pruned Action Space of the MineDojo Environment}
\label{tab:actspace_minedojo}
\begin{tabular}{ccc}
\toprule
\textbf{Index} & \textbf{Descriptions}                        & \textbf{Num of Actions} \\ \midrule
\textbf{0}     & Forward and backward                         & 3                \\
\textbf{1}     & Move left and right                          & 3                 \\
\textbf{2}     & Jump, sneak, and sprint            & 4                       \\
\textbf{3}             & Camera delta pitch/yaw ($\pm15^{\circ}$ for each action)                          & 5       \\
\textbf{4}     & Use and Attack                             & 3                \\
\bottomrule
\end{tabular}
\end{table}

\subsection{Base MBRL Methods}
\label{app:base models}

\paragraph{DreamerV2.} Unless otherwise specified, \textit{HarmonyDream (Ours)} in the experiment section refers to the HarmonyDream method based on DreamerV2 \citep{hafner2020mastering}. Details about DreamerV2 have been elaborated on in the main text, and we refer readers to Sec.~\ref{sec:overview} and \citet{hafner2019dream, hafner2020mastering}.

\paragraph{DreamerV3.}
DreamerV3 \citep{hafner2023mastering} is a general and scalable algorithm that builds upon DreamerV2. In order to master a wide range of domains with fixed hyperparameters, DreamerV3 made many changes to DreamerV2, including using symlog predictions, utilizing world model regularization by combining KL balancing and free bits, modifying the network architecture, and so forth. A main modification relevant to our method is that DreamerV3 explicitly partitions the dynamics loss in Eq.~(\ref{eq:dreamer_loss}) into a dynamics loss and a representation loss as follows:
\begin{gather}
\begin{aligned}
    \label{eq:dreamerv3_partition}
    &\text{Dynamics loss:} &&\mathcal{L}_{\text{dyn}}(\theta)=\max(1,\text{KL}\left[\mathrm{sg}(q_{\theta}(z_{t}\,|\,z_{t-1},a_{t-1}, o_{t})) \,\Vert\,p_{\theta}(\hat{z}_{t}\,|\,z_{t-1}, a_{t-1}) \right]),  \\
    &\text{Representation loss:} &&\mathcal{L}_{\text{rep}}(\theta)\,=\max(1,\text{KL}\left[q_{\theta}(z_{t}\,|\,z_{t-1},a_{t-1}, o_{t}) \,\Vert\,\mathrm{sg}(p_{\theta}(\hat{z}_{t}\,|\,z_{t-1}, a_{t-1})) \right]).  \\
\end{aligned}
\end{gather}

Since $\mathcal{L}_{\text{dyn}}(\theta)$ and $\mathcal{L}_{\text{rep}}(\theta)$ yield the same loss value, leading to identical learned coefficients, we implement Harmony DreamerV3 by recombining the two losses into $\mathcal{L}_d(\theta)$ as follows:
\begin{align}
    \label{eq:dreamerv3_dynmaics_loss}
    \mathcal{L}_d(\theta)\doteq \alpha\mathcal{L}_{\text{dyn}}(\theta)+(1-\alpha)\mathcal{L}_{\text{rep}}(\theta).
\end{align}
Here $\alpha$ is the KL balancing coefficient predefined by DreamerV3. In this way, we can use the same learning objective as Eq.~(\ref{eq:harmony_loss_with_base}) for Harmony DreamerV3.

\paragraph{DreamerPro.}
DreamerPro \citep{deng2022dreamerpro} is a reconstruction-free model-based RL method that incorporates prototypical representations in the world model learning process. 
The overall learning objective of the DreamerPro method is defined as follows:
\begin{align}
    \mathcal{L}_{\text{DreamerPro}}(\theta)=
        \mathcal{L}_{\text{SwAV}}(\theta)+\mathcal{L}_{\text{Temp}}(\theta)+\mathcal{L}_{\mathrm{R}}(\theta)+\mathcal{L}_{\mathrm{KL}}(\theta).
\end{align}
The $\mathcal{L}_{\text{SwAV}}$ term stands for prototypical representation loss used in SwAV \citep{caron2021unsupervised}, which improves prediction from an augmented view and induces useful features for static images. $\mathcal{L}_{\text{Temp}}$ stands for temporal loss that considers temporal structure and reconstructs the cluster assignment of the observation instead of the visual observation itself. As $\mathcal{L}_{\text{SwAV}}+\mathcal{L}_{\text{Temp}}$ replaces $\mathcal{L}_o$ in Eq.~(\ref{eq:dreamer_loss}), we build our Harmony DreamerPro by substituting the overall learning objective into the following:
\begin{align}
    \mathcal{L}_{\text{Harmony DreamerPro}}(\theta)=
        \sum_{i\in\{\text{SwAV},\text{Temp},\mathrm{R},\mathrm{KL}\}}\frac{1}{{\sigma_i}}\mathcal{L}_{i}(\theta) +\log{(1+\sigma_i)}.
\end{align}

\subsection{Hyperparameters}
Our proposed HarmonyDream only involves adding lightweight harmonizers, each corresponding to a single learnable parameter, and thus \textbf{does not introduce any additional hyperparameters}. For Harmony DreamerV3 and Harmony DreamerPro, we use the default hyperparameters of DreamerV3 and DreamerPro, respectively. For our HarmonyDream based on DreamerV2, we use the same set of hyperparameters as DreamerV2 \citep{hafner2020mastering}. Important hyperparameters for HarmonyDream are listed in Table \ref{tab:hyperparameter}.

\begin{table}[htbp]
\caption{Hyperparameters in our HarmonyDream based on DreamerV2. We use the same hyperparameters as DreamerV2.}
\label{tab:hyperparameter}
\begin{center}
\begin{small}
\setlength{\tabcolsep}{4.2pt}
\begin{tabular}{cll}
\toprule
                              Hyperparameter          & \multicolumn{1}{l}{Value}           \\ \midrule

                               Observation size                & $64 \times 64 \times 3$                                \\
                               Observation preprocess                & Linearly rescale from $[0, 255]$ to $[-0.5, 0.5]$  \\
                               Action Repeat & 1 for Meta-world \\
                                             & 2 for RLBench, DMCR and Natural Background DMC  \\
                               Max episode length & 500 for Meta-world, DMCR and Natural Background DMC \\ 
                                                & 200 for RLBench\\
                               Early episode termination & True for RLBench, False otherwise\\
                               Trajectory segment length $T$   &   50                              \\
                               Random exploration           & 5000 environment steps for Meta-world and RLBench \\
                                                            & 1000 environment steps for DMCR and Natural Background DMC \\
                               Replay buffer capacity        & $10^6$             \\
                               Training frequency           & Every 5 environment steps \\
                              Imagination horizon $H$ & 15 \\
                               Discount    $\gamma$   & 0.99                                  \\
                               $\lambda$-target discount & 0.95 \\
                               Entropy regularization $\eta$ & $1\times 10^{-4}$ \\
                               Batch size        & 50 for Meta-world and RLBench                         \\
                                                       & 16 for DMCR and Natural Background DMC                      \\
                                RSSM hidden size    & 1024  \\
                               World model optimizer         & Adam                                  \\
                               World model learning rate         &    $3 \times 10^{-4}$                                  \\
                               Actor optimizer         & Adam                                  \\
                               Actor learning rate       & $8 \times 10^{-5}$                                  \\
                               Critic optimizer         & Adam                                  \\
                               Critic learning rate       & $8 \times 10^{-5}$                                  \\
                               Evaluation episodes &   10   \\\bottomrule
\end{tabular}
\vspace{-10pt}
\end{small}
\end{center}
\end{table}

\subsection{Analysis Experiment Details (Fig.~\ref{fig:representation_difference} and \ref{fig:r1_r100_video_prediction_mw})}
\label{app:qualitative}
For the analysis in Sec.~\ref{sec:observaions}, namely Fig.~\ref{fig:representation_difference} and \ref{fig:r1_r100_video_prediction_mw}, we conduct our experiments on a fixed training buffer to better ablate distracting factors. We first train a separate DreamerV2 agent and use training trajectories collected during its whole training process as our fixed buffer. The fixed buffer comprises 250K environment steps and covers data from low-return to high-return trajectories \citep{levine2020offline}. We then offline train our DreamerV2 agents with different reward loss coefficients on this buffer. 
All other hyperparameters, such as training frequency, training steps, and evaluation episodes, are the same as in Table \ref{tab:hyperparameter}.

\paragraph{Details for Fig.~\ref{fig:representation_difference}} We denote the agent trained using $w_r=1$ as \textit{original weight} and trained using $w_r=100$ for Lever Pull, $w_r=10$ for Handle Pull Side and Hammer as \textit{balanced weight}.
To build the state regression dataset, first, we gather 10,000 segments of trajectories, each with a length of 50, from the evaluation episodes of both the agent trained using \textit{original weight} and the agent trained using \textit{balanced weight}. These segments are then combined into a dataset comprising 20,000 segments. This dataset is subsequently divided into a training set and a validation set at a ratio of 90\% to 10\%. Each data point in the dataset consists of a ground truth state and a predicted state representation, where the ground truth state is made up of the actual positions of task-relevant objects. We use a 4-layer MLP with a hidden size of 400 and an MSE loss to regress the representation to the ground-truth state. We report regression loss results on the validation set.

\paragraph{Details for Fig.~\ref{fig:r1_r100_video_prediction_mw}} In the Lever Pull task, the robot needs to reach the end of a lever (marked in {\color{blue}blue} in the observation) and pull it to the designated position (marked in {\color{red}red} in the observation). We utilize a trajectory 
where the default DreamerV2 with $w_r=1$ fails to lift the lever to analyze the reason behind its poor performance. Both agents use 15 frames for observation and reconstruction and predict 35 frames open-loop. We plot each image with an interval of 5 frames in Fig.~\ref{fig:r1_r100_video_prediction_mw}.

\subsection{Computational Resources}
We implement our HarmonyDream based on DreamerV2 using PyTorch \citep{paszke2019pytorch}. Training is conducted with automatic mixed precision \citep{micikevicius2018mixed} on Meta-world and RLBench and full precision on DMCR. In terms of training time, it takes $\sim$24 hours for each run of Meta-world experiments over 250K environment steps, $\sim$24 hours for RLBench over 500K environment steps, and $\sim$23 hours for DMCR over 1M environment steps, respectively. The lightweight harmonizers introduced by HarmonyDream do not affect the training time. In terms of memory usage, Meta-world and RLBench experiments require $\sim$10GB GPU memory, and DMCR requires $\sim$5GB GPU memory, thus, the experiments can be done using typical 12GB GPUs.

\section{Extended Discussions}

\subsection{Differences with DreamerV3}
\label{app:difference_dreamerv3}
When we started this work, DreamerV3 had not been released. Thus, we primarily conduct experiments based on DreamerV2, as mentioned in the main paper. We state here that the modifications introduced by DreamerV3 do not fully address the problem of task domination inside world models, which is the problem HarmonyDream intends to solve. As shown in Appendix \ref{app:atari_100k_experiments} and \ref{app:dreamerv3_metaworld}, HarmonyDream applied to DreamerV3 can further unleash the potentials of this base method.

There are mainly two changes of DreamerV3 relevant to improving world model learning: KL balancing and symlog predictions. We have already shown in Appendix \ref{app:base models} that KL balancing is orthogonal to our method and that we can easily incorporate this modification into our approach. On the other hand, symlog predictions also do not solve our problem of seeking a balance between reward modeling and observation modeling. First of all, the symlog transformation only shrinks extremely large values but is unable to rescale various values into exactly the same magnitude, while our harmonious loss properly addresses this by dynamically approximating the scales of the values. More importantly, the primary reason why $L_r$ has a significantly smaller loss scale is the difference in dimension: as we have stated in Sec~\ref{sec:observaions}, the observation loss $L_o$ usually aggregates $H\times W\times  C$ dimensions, while the reward loss $L_r$ is derived from only a scalar. In summary, using symlog predictions as DreamerV3 only mitigates the problem of differing per-dimension scales (typically across environment domains) by a static transformation, while our method aims to dynamically balance the overall loss scales across tasks in world model learning, considering together per-dimension scales, dimensions, and training dynamics.

In practice, DreamerV3 uses twohot symlog predictions for the reward predictor to replace the MSE loss in DreamerV2. This approach increases the scale of the reward loss, but is insufficient to mitigate the domination of the image loss. We observe that the reward loss in DreamerV3 is still significantly smaller than the observation loss, especially for visually demanding domains such as RLBench, where the reward loss is still two orders of magnitude smaller.

\subsection{Comparisons with Multi-task Learning Methods}

\label{app:multi-task-baseline}
In this paper, we understand world model learning from a multi-task or multi-objective view.
Methods in the field of multi-task learning or multi-objective learning can be roughly categorized into loss-based and gradient-based. Since gradient-based methods mainly address the problem of gradient conflicts \citep{yu2020gradient, liu2021conflictaverse}, which is not the main case in world model learning, we focus our discussion on loss-based methods, which assigns different weights to task losses by various criteria. We choose the following methods as our baselines to discuss differences and conduct comparison experiments. The experiment results can be found in Fig.~\ref{fig:multi-task-baseline} of the main paper.

\vspace{-5pt}
\paragraph{Uncertainty Weighting (UW, \citet{kendall2018multitask})} balances tasks with different scales of targets, which is measured as uncertainty of outputs. 
As pointed out in Section \ref{sec:overview}, in world model learning, observation loss $\mathcal{L}_o(\theta)=-\log p_\theta\left(o_t \mid z_t\right)=-\sum_{h,w,c} \log p_{\theta}(o_{t}^{(h,w,c)}\,|\,z_{t})$ and reward loss $\mathcal{L}_r(\theta)=-\log p_\theta\left(r_t \mid z_t\right)$ differs not only in scales but also in dimensions.
To implement UW, we opt for depicting the uncertainty of each pixel. By assuming all pixel values share a common standard deviation $\sigma_o$ for Gaussian distributions, the uncertainty-weighted image loss takes the following form:
$
    \mathcal{L}(\theta, \sigma_{o}) = \sum_{h,w,c} (\hat{o}_t^{(h,w,c)} - o_t^{(h,w,c)})^2 / 2{\sigma_o} + \log{\sigma_o} = {\sigma_o}^{-1}\mathcal{L}_{o}(\theta) + HWC \log{\sigma_o}.
$
A detailed explanation of the differences between our harmonious loss and UW is provided in the discussion section in Section \ref{sec:method}.

\vspace{-5pt}
\paragraph{Dynamics Weight Average (DWA, \citet{liu2019endtoend})} balances tasks according to their learning progress, illustrating the various task difficulties. However, in world model learning, since the data in the replay buffer is growing and non-stationary, the relative descending rate of losses may not accurately measure task difficulties and learning progress.

\vspace{-5pt}
\paragraph{NashMTL \citep{navon2022multitask}} is the most similar to our method, whose optimization direction has balanced projections of individual gradient directions. However, its implementation is far more complex than our method, as it introduces an optimization procedure to determine loss weights on each iteration. In our experiments, we also find this optimization is prone to deteriorate to produce near-zero weights without careful tuning of optimization parameters.

In Fig~\ref{fig:multi-task-baseline}, we compare against the multi-task methods we mentioned above. Experiments are conducted on \textit{Lever Pull} from Meta-world, \textit{Push Button} from RLBench, and \textit{Cheetah Run} from DMCR, respectively. Our method is the most effective among multi-task methods and has the advantage of simplicity. Although NashMTL produces similar results on the Lever Pull task, it outputs extreme loss weights on the other two tasks, which accounts for its low performance. Our HarmonyDream, on the other hand, uses a rectified loss that effectively mitigates extremely large loss weights.

\newpage
\section{Extended Experiment Results}
\label{app:additional experiments}

\subsection{Atari 100K Experiments}
\label{app:atari_100k_experiments}
\begin{figure*}[!h]
    \centering
    \vspace{-10pt}
    \includegraphics[width=\textwidth]{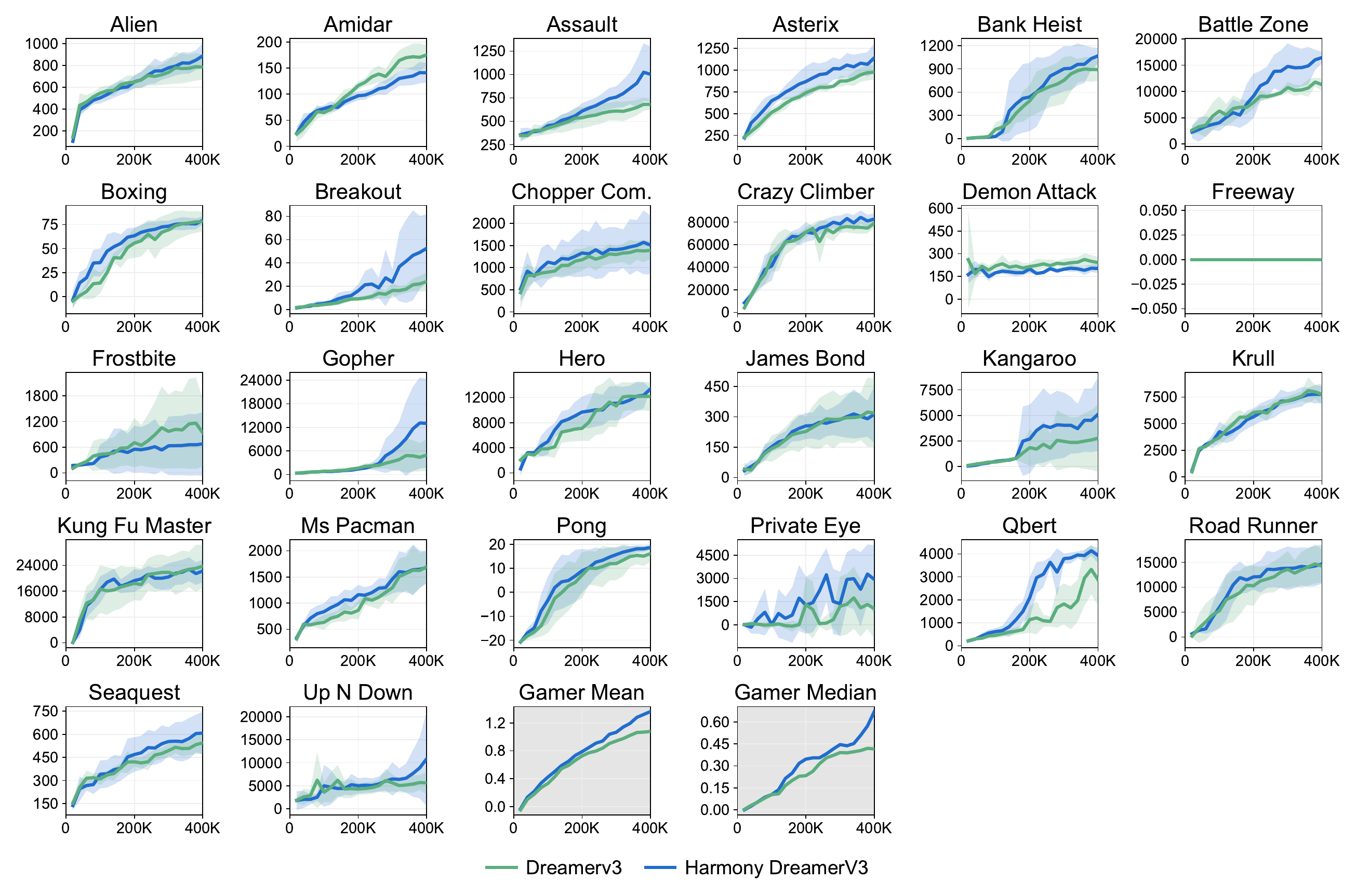}
    \caption{Atari learning curves of DreamerV3 (reproduced) and Harmony DreamerV3 with a budget of 400K frames, amounting to 100K policy steps.}
    \label{fig:atari_learning_curve}
\end{figure*}

We based our implementation of HarmonyDream applied to DreamerV3 (denoted as \textit{Harmony DreamerV3}) on the official DreamerV3 codebase\footnote{We use this version of the DreamerV3 codebase: \url{https://github.com/danijar/dreamerv3/tree/8fa35f}. We notice that several changes have made to this codebase subsequent to our paper's initial release in February 2024.}. To ensure the fairness and quality of our results, we also reproduced DreamerV3 results using the official code and configurations. 
Fig~\ref{fig:atari_learning_curve} shows Atari learning curves of the reproduced DreamerV3 and our Harmony DreamerV3 on all 26 environments.  
Note here that our learning curves are plotted using \textbf{evaluation scores}, rather than averaged training scores as in DreamerV3, which may account for part of the differences between our curves and that reported by \citet{hafner2023mastering}. Both DreamerV3 and our Harmony DreamerV3 are evaluated for 100 episodes every 20K environment steps. In each curve, the solid line represents the average evaluation score across 5 seeds, while the shaded region indicates the standard deviation. This is consistent with the figure representation in DreamerV3.

Table~\ref{tab:atari_results} shows the mean score and aggregated human normalized scores of our Harmony DreamerV3 on Atari tasks, compared to other methods. The scores in the \textit{SimPLe, TWM, IRIS}, and \textit{DreamerV3 (Original)} columns correspond to the scores reported in their papers, respectively. The \textit{DreamerV3 (Reproduced)} column contains scores reproduced using the official codebase. The reproduced results exhibit performance comparable to the reported results. The slight discrepancy in the human-normalized score is primarily attributed to the subpar performance in the Crazy Climber game. Our Harmony DreamerV3 significantly improves upon the base method’s performance. It either matches or surpasses DreamerV3 in 23 of the 26 tested environments, thereby setting a new state-of-the-art benchmark with a human mean score of 136.5\%. It’s noteworthy that this enhancement is achieved without the addition of any hyperparameters or alterations to any network structures. By harmonizing tasks in world model learning, we fully exploit the inherent potential of our base model, further highlighting the value of our work.

\begin{table}[h]
    \centering
    \caption{Mean scores on the Atari 100K benchmark per game as well as the aggregated human normalized mean and median. Bold numbers indicate scores within 5\% of the best.}
    \vspace{10pt}
    \label{tab:atari_results}
    \small
    \begin{tabular}{lrrrrrrrr}
    \toprule
        Game & Random & Human & \makecell{SimPLe \\ \citeyearpar{kaiser2019model}}  &  \makecell{TWM \\ \citeyearpar{robine2023transformerbased}} & \makecell{IRIS \\ \citeyearpar{micheli2022transformers}} & \makecell{DreamerV3\\(Original)} & \makecell{DreamerV3\\(Reproduced)} & \makecell{Harmony\\DreamerV3} \\ \midrule
        Alien & 228 & 7128 & 617 & 675 & 420 & \textbf{959} & 786  & 890  \\ 
        Amidar & 6 & 1720 & 74 & 122 & 143 & 139 & \textbf{175}  & 141  \\ 
        Assault & 222 & 742 & 527 & 683 & \textbf{1524} & 706 & 680  & 1003  \\ 
        Asterix & 210 & 8503 & \textbf{1128} & \textbf{1117} & 854 & 932 & 974  & \textbf{1140}  \\ 
        Bank Heist & 14 & 753 & 34 & 467 & 53 & 649 & 894  & \textbf{1069}  \\ 
        Battle Zone & 2360 & 37188 & 4031 & 5068 & 13074 & 12250 & 11314  & \textbf{16456}  \\ 
        Boxing & 0 & 12 & 8 & \textbf{78} & 70 & \textbf{78} & \textbf{78}  & \textbf{80}  \\ 
        Breakout & 2 & 30 & 16 & 20 & \textbf{84} & 31 & 24  & 53  \\ 
        Chopper Com. & 811 & 7388 & 979 & \textbf{1697} & 1565 & 420 & 1390  & 1510  \\ 
        Crazy Climber & 10780 & 35829 & 62584 & 71820 & 59234 & \textbf{97190} & 78969  & 82739  \\ 
        Demon Attack & 152 & 1971 & 208 & 350 & \textbf{2034} & 303 & 241  & 203  \\ 
        Freeway & 0 & 30 & 17 & 24 & \textbf{31} & 0 & 0  & 0  \\ 
        Frostbite & 65 & 4335 & 237 & \textbf{1476} & 259 & 909 & 939  & 679  \\ 
        Gopher & 258 & 2412 & 597 & 1675 & 2236 & 3730 & 4936  & \textbf{13043}  \\ 
        Hero & 1027 & 30826 & 2657 & 7254 & 7037 & 11161 & 12160  & \textbf{13378}  \\ 
        James Bond & 29 & 303 & 101 & 362 & \textbf{463} & \textbf{445} & 318  & 317  \\ 
        Kangaroo & 52 & 3035 & 51 & 1240 & 838 & 4098 & 2773  & \textbf{5118}  \\ 
        Krull & 1598 & 2666 & 2205 & 6349 & 6616 & \textbf{7782} & \textbf{7764}  & \textbf{7754}  \\ 
        Kung Fu Master & 258 & 22736 & 14862 & \textbf{24555} & 21760 & 21420 & \textbf{23753}  & 22274  \\ 
        Ms Pacman & 307 & 6952 & 1480 & 1588 & 999 & 1327 & \textbf{1696}  & \textbf{1681}  \\ 
        Pong & -21 & 15 & 13 & \textbf{19} & 15 & \textbf{18} & \textbf{18}  & \textbf{19}  \\ 
        Private Eye & 25 & 69571 & 35 & 87 & 100 & 882 & 1036  & \textbf{2932}  \\ 
        Qbert & 164 & 13455 & 1289 & 3331 & 746 & 3405 & 2872  & \textbf{3933}  \\ 
        Road Runner & 12 & 7845 & 5641 & 9109 & 9615 & \textbf{15565} & 14248  & 14646  \\ 
        Seaquest & 68 & 42055 & 683 & \textbf{774} & 661 & 618 & 544  & 665  \\ 
        Up N Down & 533 & 11693 & 3350 & \textbf{15982} & 3546 & 7667 & 5636  & 10874 \\ \midrule
        Human Mean & 0\% & 100\% & 33\% & 96\% & 105\% & 112\% & 108\% & \textbf{136.5\%} \\ 
        Human Median & 0\% & 100\% & 13\% & 51\% & 29\% & 49\% & 42\% & \textbf{67.1\%} \\ \bottomrule
    \end{tabular}
\end{table}

\subsection{DeepMind Control Suite Experiments}
\begin{figure*}[h]
    \centering
    \includegraphics[width=\textwidth]{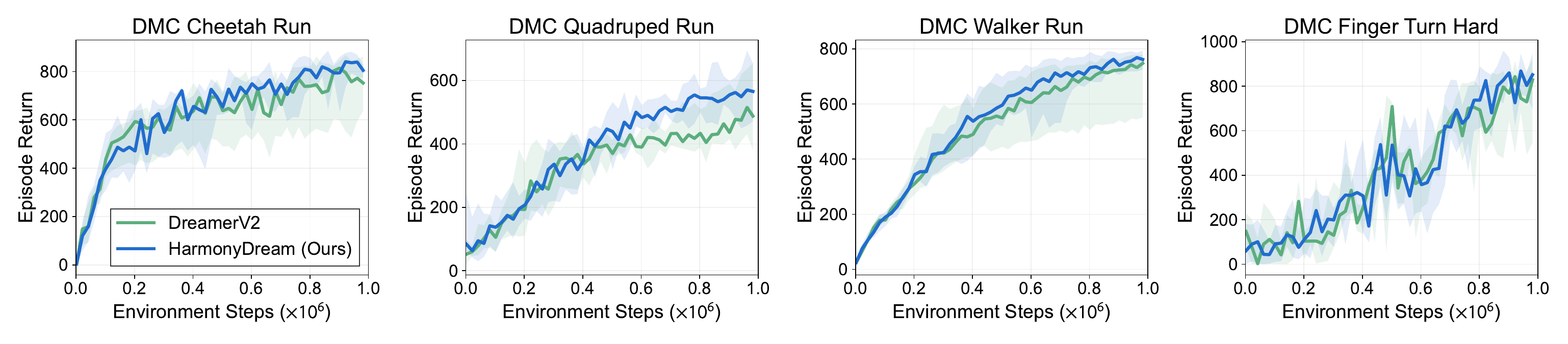}
    {
    \vspace{-15pt}
    \caption{Learning curves of HarmonyDream and DreamerV2 on the DMC domain.}
    \label{fig:dmc_result}
    }
    \vspace{-10pt}
\end{figure*}

The DeepMind Control Suite (DMC, \citet{tassa2018deepmind}) is a widely used benchmark for visual locomotion. We have conducted additional experiments on four tasks: Cheetah Run, Quadruped Run, Walker Run, and Finger Turn Hard. In Fig.~\ref{fig:dmc_result}, we present comparisons between our HarmonyDream and the base DreamerV2. 
We note that the performance of relatively easy DMC tasks has been almost saturated by recent literature \citep{yarats2021improving, hafner2020mastering}, and we suppose that in this domain, current limitations of model-based methods are not rooted in the world model, but rather in behavior learning \citep{hafner2023mastering}, which falls outside the scope of our method and contributions. 
Nevertheless, our HarmonyDream is still able to obtain a noticeable gain in performance in the more difficult Quadruped Run task.

\subsection{Ablation Study on Rectified Harmonious Loss}

\begin{figure}[!h]
    \vspace{-5pt}
    \centering
    \begin{subfigure}[t]{0.27\textwidth}
        \centering
        \includegraphics[width=\textwidth]{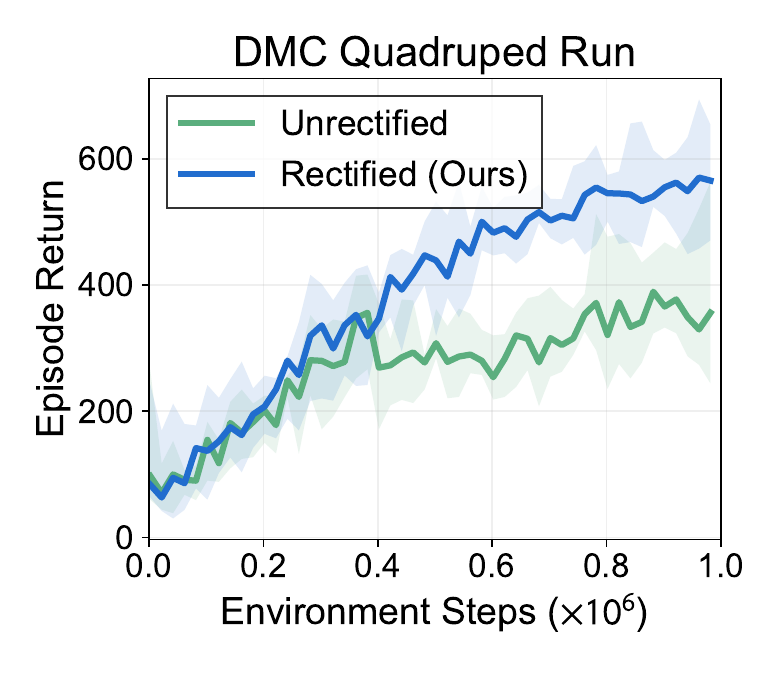}
        {\vspace{-20pt}
        \caption{Learning curves.}
        \label{fig:unrectified_HarmonyDream_dmc_learning_curve}}
    \end{subfigure}
    \hspace{-3pt}
    \begin{subfigure}[t]{0.27\textwidth}
        \centering
        \includegraphics[width=\textwidth]{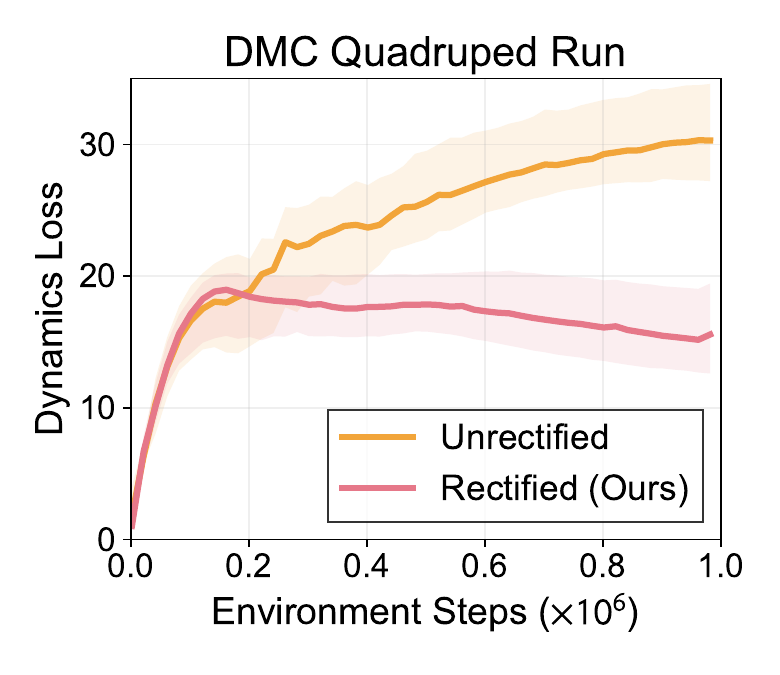}
        {\vspace{-20pt}
        \caption{Dynamics loss.}
        \label{fig:unrectified_HarmonyDream_dmc_kl_loss}}
    \end{subfigure}%
    \hspace{-3pt}
    \begin{subfigure}[t]{0.27\textwidth}
        \centering
        \includegraphics[width=\textwidth]{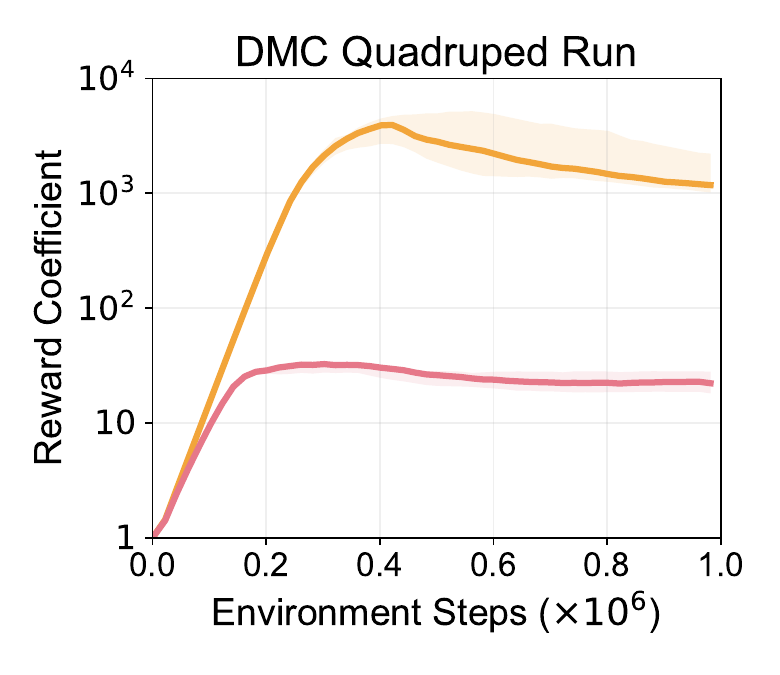}
        {\vspace{-20pt}
        \caption{Reward coefficient.}
        \label{fig:unrectified_HarmonyDream_dmc_reward_coeff}}
    \end{subfigure}%
    {\vspace{-5pt}
    \caption{Training curves for \textit{Unrectified HarmonyDream} (denoted as Unrectified) using Eq.~(\ref{eq:harmony_loss}) on the DMC Quadruped Run task, in comparison with our HarmonyDream (denoted as Rectified). }
    \label{fig:unrectified_HarmonyDream_dmc}}
    \vspace{-5pt}
\end{figure}

\begin{figure}[!h]
    \centering
    \includegraphics[width=0.8\textwidth]{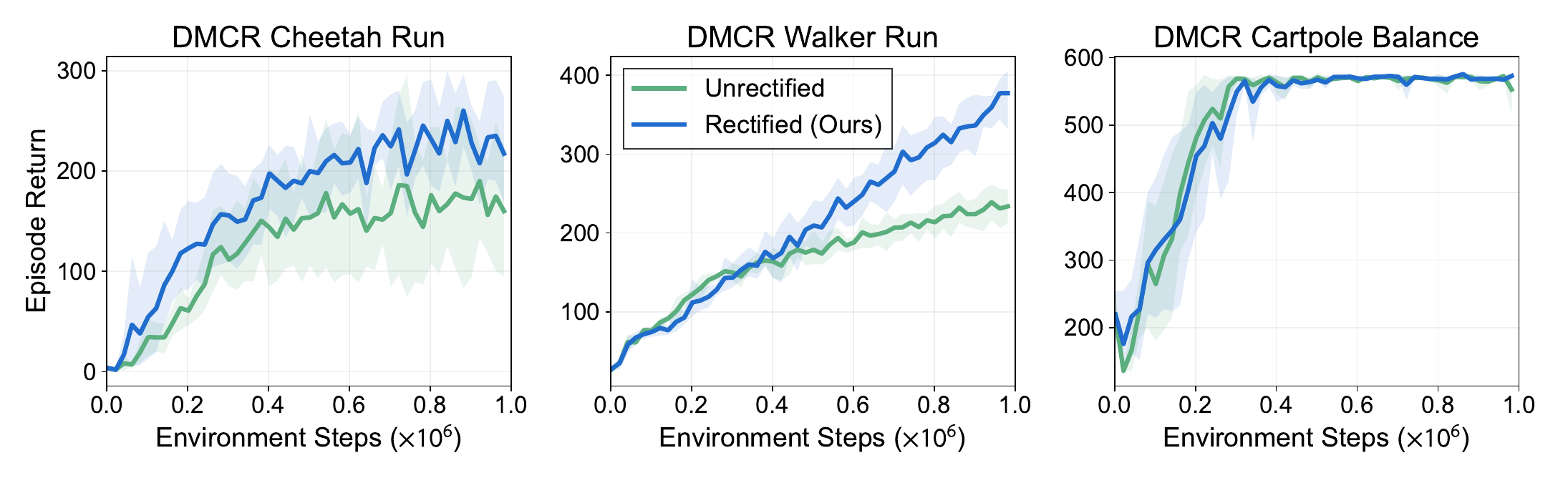}
    {\vspace{-10pt}
    \caption{Learning curves for \textit{Unrectified HarmonyDream} (denoted as Unrectified) using Eq.~(\ref{eq:harmony_loss}) on the DMCR domain, in comparison with our HarmonyDream (denoted as Rectified).}
    \label{fig:unrectified_HarmonyDream_dmcr}}
    \vspace{-10pt}
\end{figure}

In Sec.~\ref{sec:method}, we have already presented a detailed explanation on the necessity of our \textit{rectified harmonious loss}, changing the regularization term from $\log \sigma_i$ in Eq.~(\ref{eq:harmony_loss}) to $\log (1+\sigma_i)$ in Eq.~(\ref{eq:harmony_loss_with_base}). Here, we present experimental results to support our claim. We use \textit{Unrectified} to note our method trained using the objective in Eq.~(\ref{eq:harmony_loss}), and \textit{Rectified (Ours)} to note our method trained using Eq.~(\ref{eq:harmony_loss_with_base}). As shown in Fig.~\ref{fig:unrectified_HarmonyDream_dmc} and Fig.~\ref{fig:unrectified_HarmonyDream_dmcr}, the excessively large reward coefficient (Fig.~\ref{fig:unrectified_HarmonyDream_dmc_reward_coeff}) for \textit{Unrectified} can lead to a divergence in the dynamics loss (Fig.~\ref{fig:unrectified_HarmonyDream_dmc_kl_loss}), which in turn negatively impacts performance (Fig.~\ref{fig:unrectified_HarmonyDream_dmc_learning_curve} and Fig.~\ref{fig:unrectified_HarmonyDream_dmcr}).

\subsection{Ablation Study on Adjusting Dynamics Loss Weight $w_d$}

\begin{figure}[ht]
    \centering
    \includegraphics[width=0.8\textwidth]{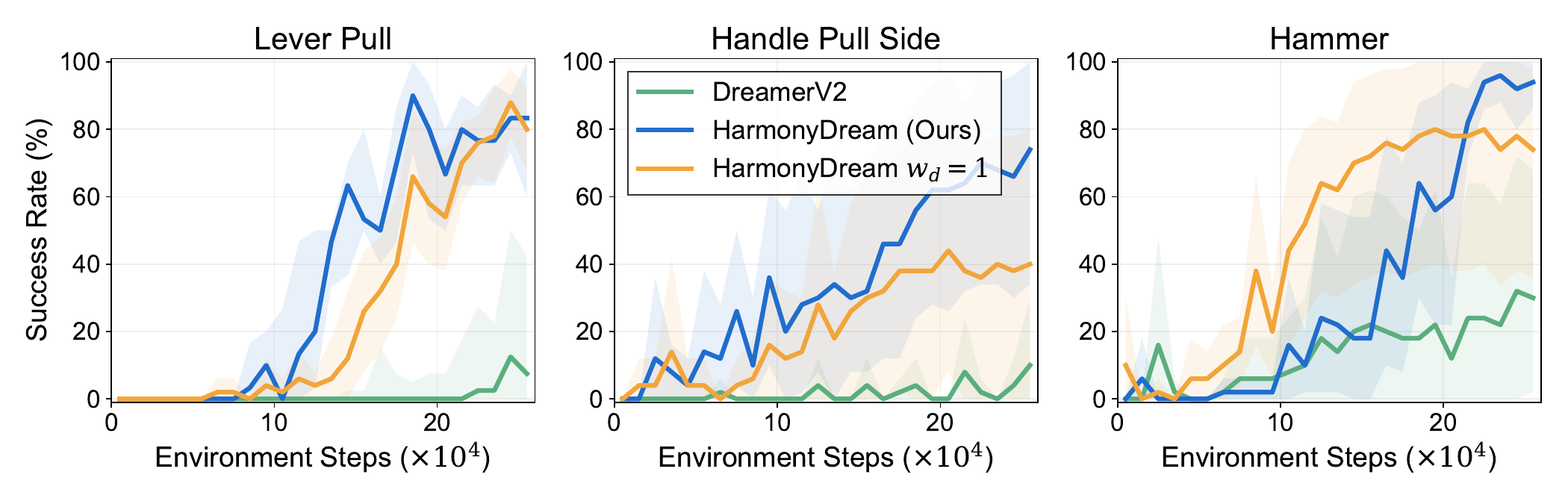}
    {\vspace{-10pt}
    \caption{Ablation on adjusting $w_d$ in HarmonyDream. }
    \label{fig:adjust_kl_ablation}}
    \vspace{-10pt}
\end{figure}

Manually tuning the dynamics loss coefficient $w_d$ (e.g. $w_d=0.1$) is common in MBRL methods \citep{hafner2020mastering, hafner2023mastering, seo2022masked, seo2022reinforcement}. We note that our HarmonyDream differs from these previous approaches as we treat the different losses in a multi-task view and balance loss scales between them, while previous approaches see $w_d$ simply as a hyperparameter. Fig.~\ref{fig:adjust_kl_ablation} shows a comparison between fixing $w_d$ to $1$ in HarmonyDream (denoted as \textit{HarmonyDream $w_d=1$}) and using $\sigma_d$ to balance $w_d$ (denoted as \textit{HarmonyDream (Ours)}), where our proposed HarmonyDream performs slightly better than the one fixing $w_d$, and both methods outperform DreamerV2 by a clear margin. 
This result highlights the importance of harmonizing two different modeling tasks in world models, instead of only tuning on the shared dynamics part of them.

\subsection{Comparison to Tuned Weights}
\begin{figure*}[h]
    \centering
    \begin{subfigure}[t]{0.8\textwidth}
        \centering
        \includegraphics[width=\textwidth]{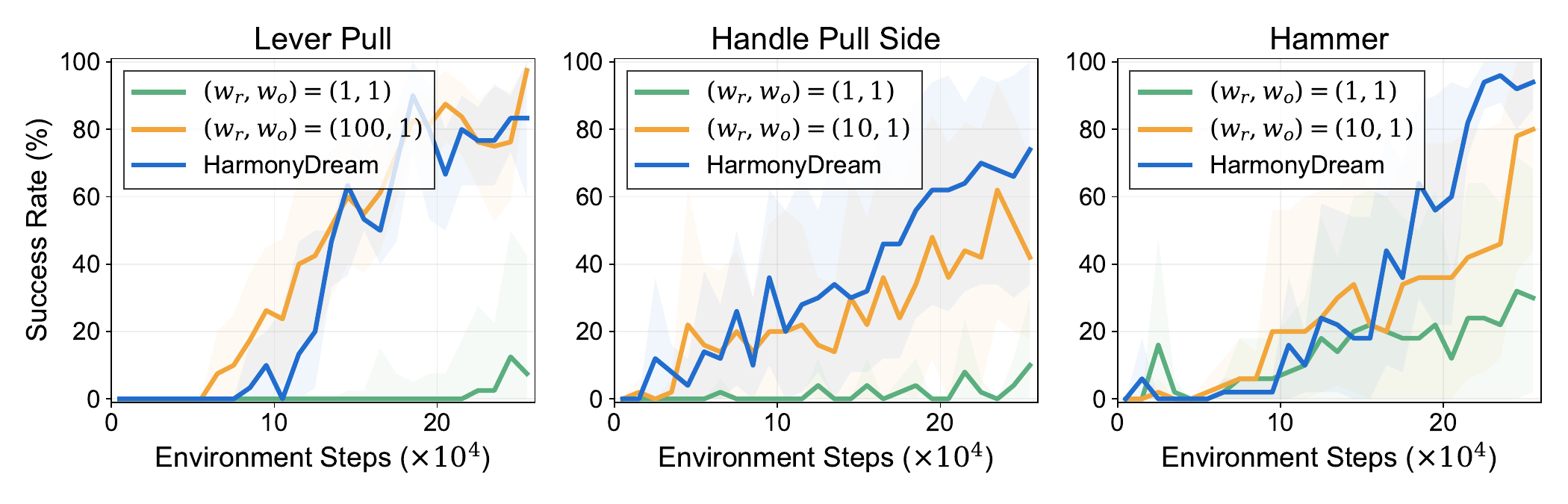}
        \caption{Meta-world.}
    \end{subfigure}
    
    \begin{subfigure}[t]{0.81\textwidth}
        \centering
        \includegraphics[width=\textwidth]{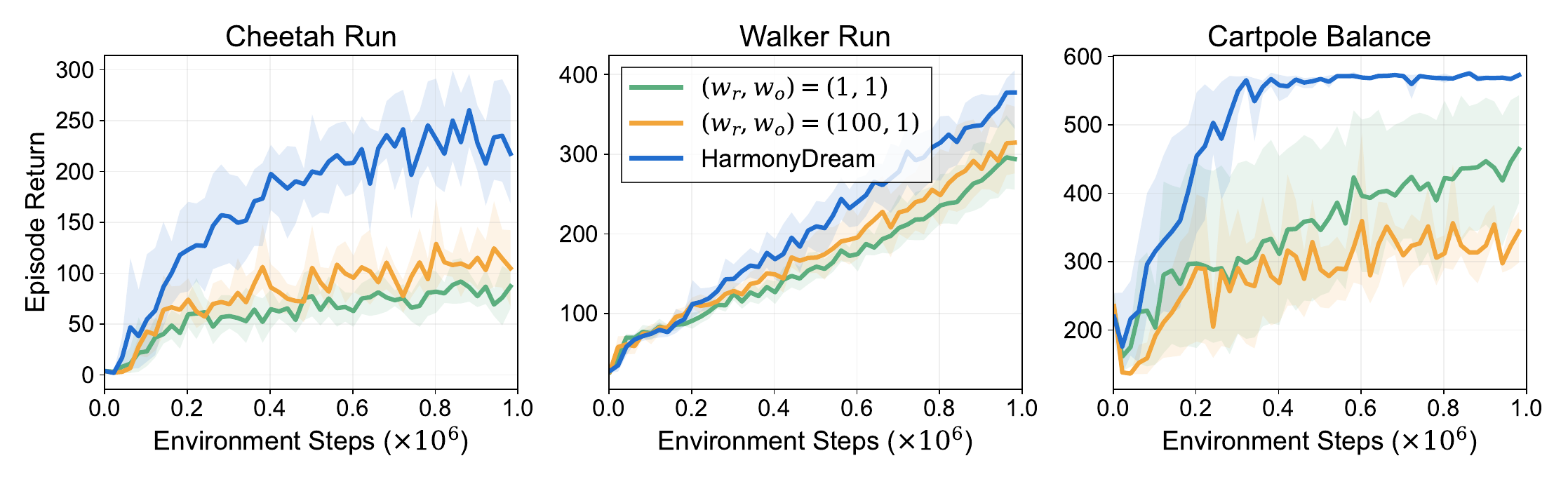}
        \caption{DMCR.}
    \end{subfigure}%
    \caption{Learning curves of HarmonyDream compared to tuned weights on Meta-world and DMCR.}
    \label{fig:tuned_weights_comparison}
    \vspace{-10pt}
\end{figure*}

We present a direct comparison between our HarmonyDream and manually tuned weights for DreamerV2. For the Meta-world domain, we plot the tuned better results from $w_r\in\{10,100\}, w_o=1$. For the DMCR domain, we plot tuned results using $w_r=100, w_o=1$. Results in Fig.~\ref{fig:tuned_weights_comparison} show that our HarmonyDream outperforms manually tuned weights in most tasks, which adds to the value of our method.

\subsection{Extended Results of Harmony DreamerV3 on Meta-world}
\label{app:dreamerv3_metaworld}
\begin{figure*}[htbp]
    \centering
    \includegraphics[width=\textwidth]{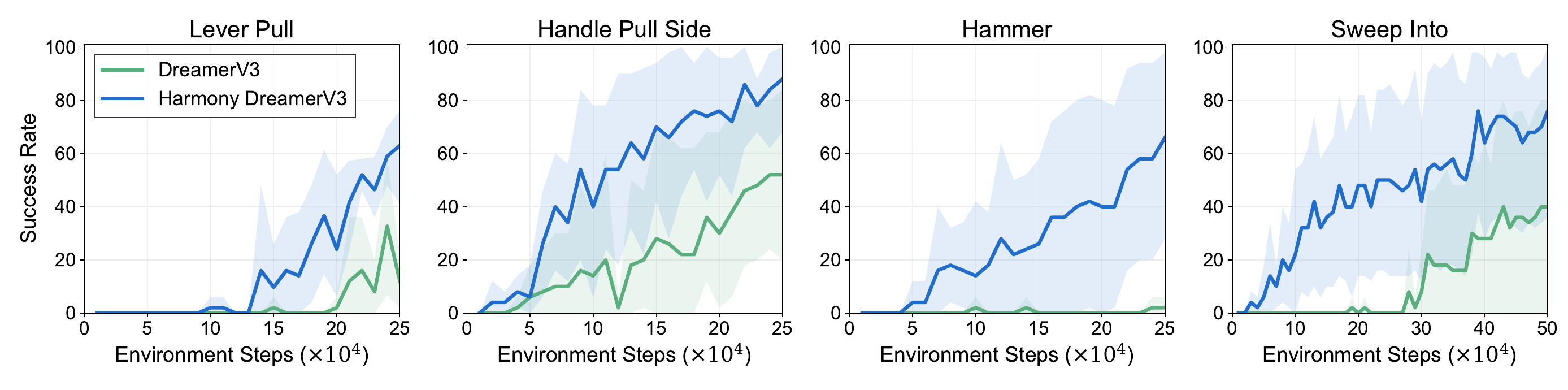}
    \caption{Detailed results of Harmony DreamerV3 on Meta-world.}
    \label{fig:additional_method_generality_results}
\end{figure*}

In Fig.~\ref{fig:model_generality} of the main paper, we have presented the aggregated results of our HarmonyDream generalized to DreamerV3 (referred to as \textit{Harmony DreamerV3}), on three Meta-world tasks: Lever Pull, Handle Pull Side, and Hammer. Here in Fig.~\ref{fig:additional_method_generality_results}, we provide individual results of these three tasks, along with the results of an additional task, Sweep Into. Our approach consistently improves the sample efficiency of our base method, proving excellent generality.

\subsection{Additional Results of Harmony DreamerV3 on Crafter}
\label{app:dreamerv3_crafter_minecraft}

\begin{figure*}[!h]
    \centering
    \includegraphics[width=0.3\textwidth]{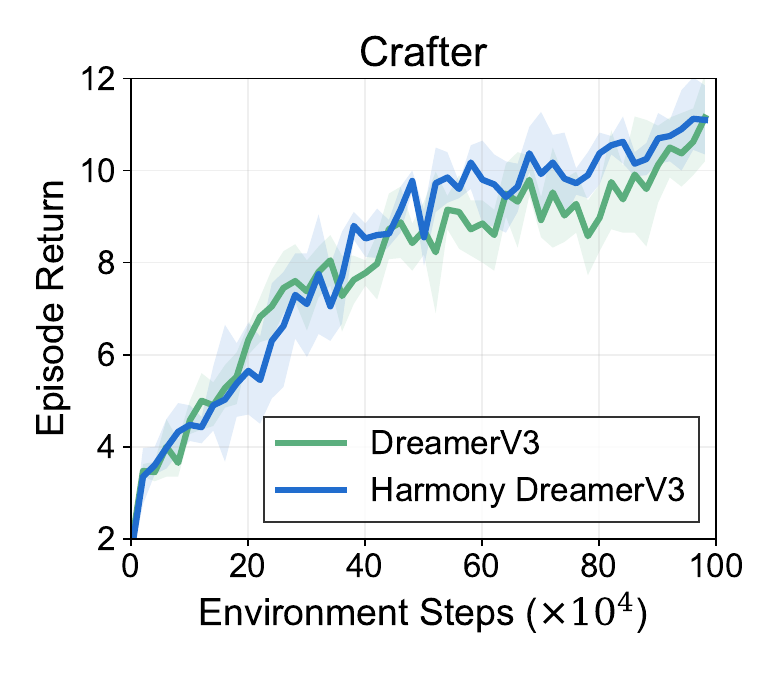}
    {\vspace{-15pt}
    \caption{Results of Harmony DreamerV3 on Crafter.}
    \label{fig:dreamerv3_crafter}}
    \vspace{-10pt}
\end{figure*}

Crafter \citep{hafner2022benchmarking} is a 2D open-world survival game benchmark where the agent needs to learn multiple skills within a single environment. High rewards in this benchmark demand robust generalization and representation capabilities from the agent. However, our method is not typically effective in the Crafter domain, which is characterized by clear observations and distinct target objects. As a result, Harmony DreamerV3 marginally outperforms DreamerV3, as shown in Fig.~\ref{fig:dreamerv3_crafter}.

\subsection{Extended Results of Implicit MBRL Methods}
\begin{figure}[!ht]
    \vspace{-5pt}
    \centering
    \begin{subfigure}[t]{0.8\textwidth}
        \centering
        \includegraphics[width=\textwidth]{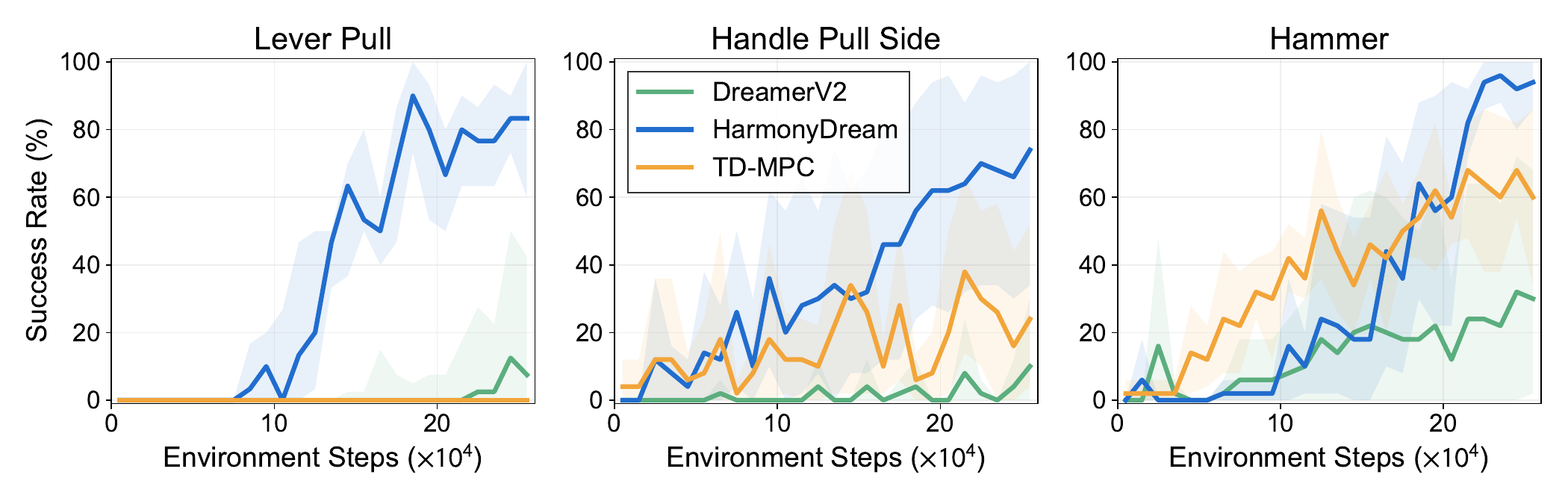}
        {\vspace{-20pt}
        \caption{Meta-world.}}
    \end{subfigure}
    \begin{subfigure}[t]{0.81\textwidth}
        \centering
        \includegraphics[width=\textwidth]{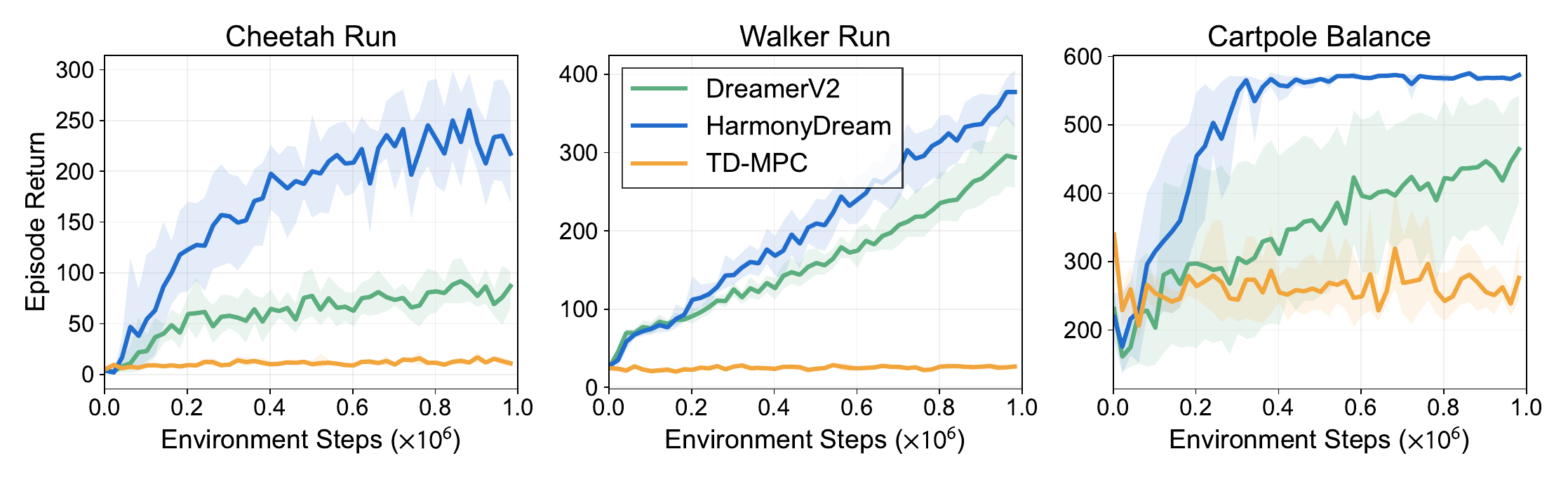}
        {\vspace{-20pt}
        \caption{DMCR.}}
    \end{subfigure}
    {\vspace{-5pt}
    \caption{Learning curves of TD-MPC.}
    \label{fig:tdmpc_full_result}}
    \vspace{-10pt}
\end{figure}

We observe that the performance of TD-MPC \citep{hansen2022temporal} is fairly low compared to our HarmonyDream. Due to our limited computational resources, we only conduct experiments on the Meta-world and DMCR domain. The Meta-world result in Fig.~\ref{fig:tdmpc_dmcr_aggregated} aggregates over three tasks: Lever Pull, Handle Pull Side, and Hammer, which are the same three tasks as in Fig.~\ref{fig:model_generality}. Full TD-MPC results in Fig.~\ref{fig:tdmpc_full_result} show that TD-MPC is unable to learn a meaningful policy in some tasks.

\subsection{Comparison with Denoised MDP}
\label{app:comparison_denoised_mdp}
\begin{figure}[ht]
    \vspace{-5pt}
    \centering
    \begin{subfigure}[t]{0.53\textwidth}
        \centering
        \includegraphics[width=\textwidth]{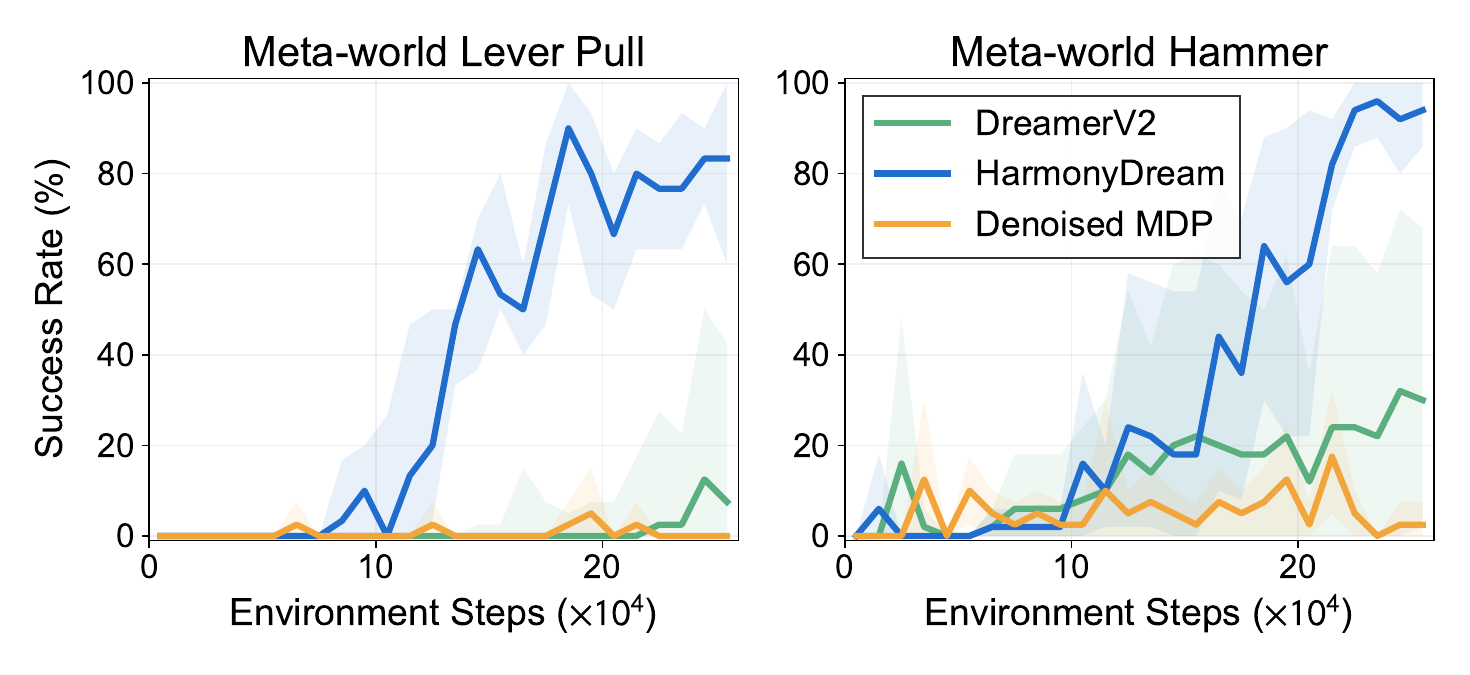}
    \end{subfigure}
    \begin{subfigure}[t]{0.28\textwidth}
        \centering
        \includegraphics[width=\textwidth]{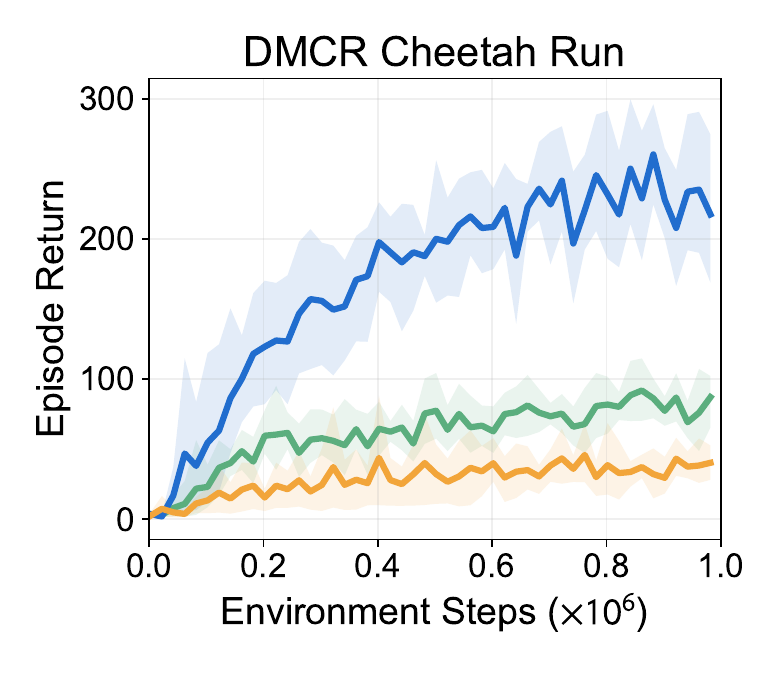}
    \end{subfigure}
    {\vspace{-10pt}
    \caption{Comparison of HarmonyDream with Denoised MDP.}
    \label{fig:comparison_denoised_mdp_results}}
\end{figure}

HarmonyDream shares a similar point with Denoised MDP \citep{wang2022denoisedmdps} in enhancing task-centric representations. 
However, the two approaches are orthogonal. 
In Fig.~\ref{fig:comparison_denoised_mdp_results}, we show a comparison of our method to Denoised MDP. Denoised MDP performs information decomposition by changing the MDP transition structure and utilizing the reward as a guide to separate task-relevant information. However, since Denoised MDP does not modify the weight for the reward modeling task, the observation modeling task can still dominate the learning process. Consequently, the training signals from the reward modeling task may be inadequate to guide decomposition. It's also worth noting that Denoised MDP only added noise distractors to task-irrelevant factors in their DMC experiments. On the other hand, the benchmark adopted in our experiments, DMCR, adds visual distractors to both task-irrelevant and task-relevant factors, such as the color of the body and floor, which adds complexity to both factors and results in more challenging tasks. These two reasons above can account for the low performance of Denoised MDP in our benchmarks.

\vspace{-10pt}

\subsection{Comparison with RePo}
\label{app:comparison_repo}
\begin{figure}[!ht]
    \vspace{-5pt}
    \centering
    \begin{subfigure}[t]{0.8\textwidth}
        \centering
        \includegraphics[width=\textwidth]{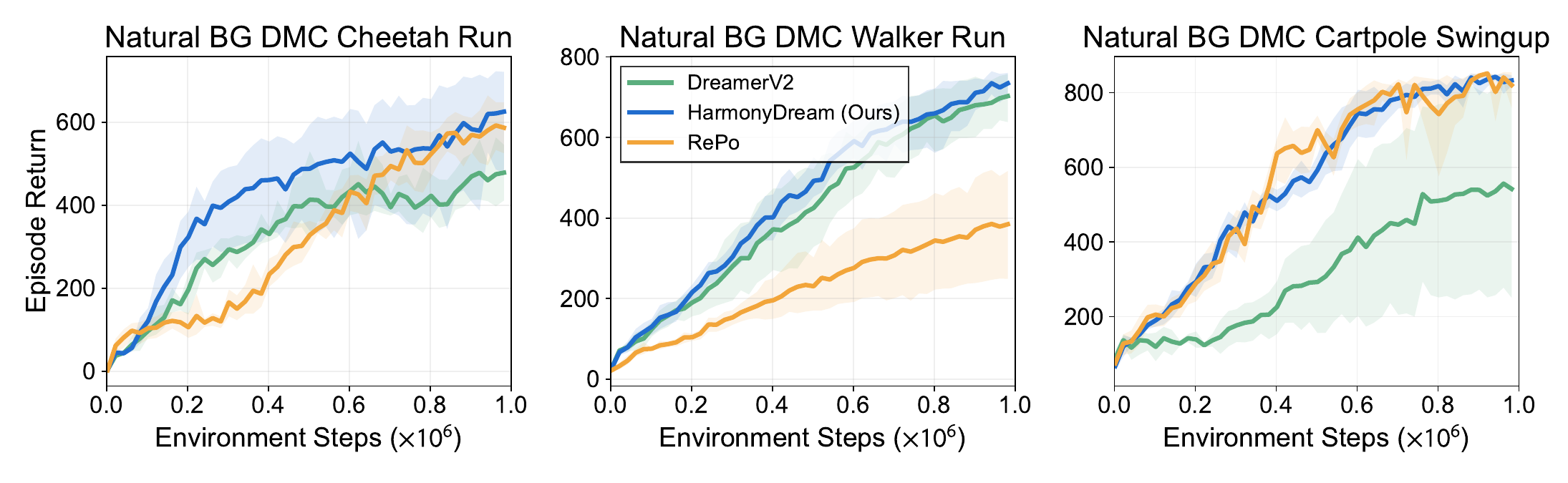}
        {\vspace{-20pt}
        \caption{Natural background (BG) DMC.}}
    \end{subfigure}
    \begin{subfigure}[t]{0.81\textwidth}
        \centering
        \includegraphics[width=\textwidth]{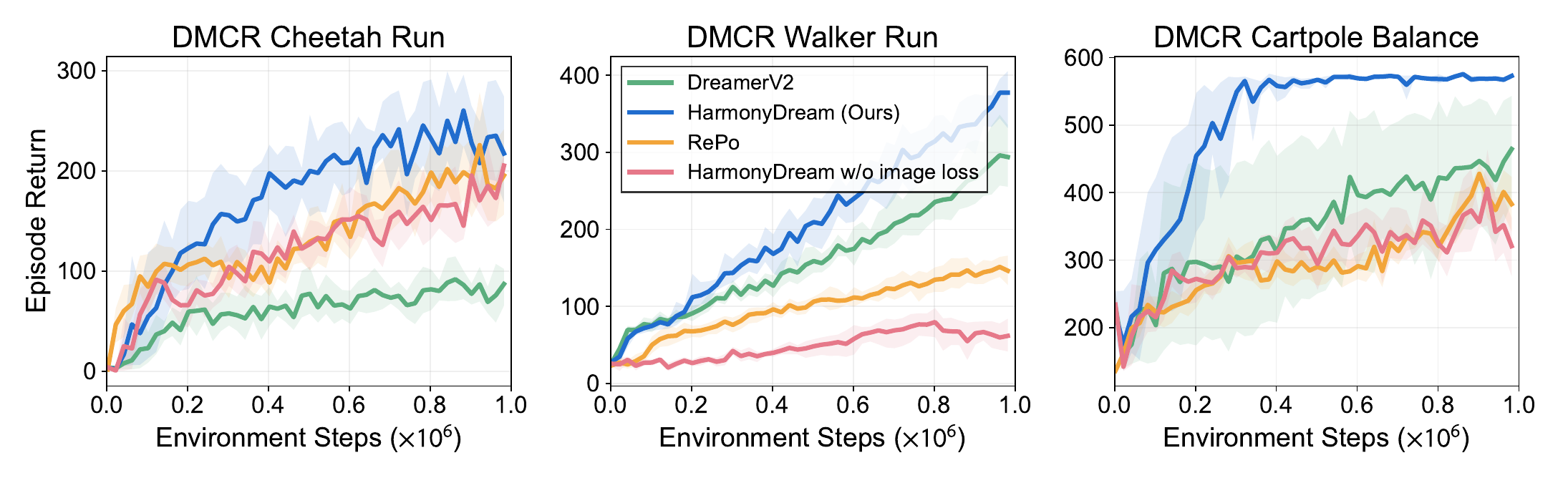}
        {\vspace{-20pt}
        \caption{DMCR.}}
    \end{subfigure}
    {\vspace{-5pt}
    \caption{Comparison of HarmonyDream with RePo.}
    \label{fig:repo_full_results}}
    \vspace{-10pt}
\end{figure}

RePo \citep{zhu2023repo} is a modification on DreamerV2 that removes the observation reconstruction loss while introducing a dynamically adjusted coefficient of dynamics loss. As shown in Fig.~\ref{fig:repo_full_results}, our HarmonyDream shows a higher sample efficiency compared to RePo on both natural background DMC and DMCR. It is notable that RePo takes a similar form as HarmonyDream without observation loss (i.e. fixing $w_o=0$). While the adjusted coefficient of RePo does not guarantee uniform loss scales, we observe in our experiments that it, in effect, makes dynamics loss and reward loss have more similar scales. We demonstrate on the DMCR domain that the two share similar learning curves, which to some extent enhances our Finding 3, that learning signals from rewards alone is inadequate for sample-efficient learning due to limited representation learning capability. We also note that RePo still needs to carefully tune a crucial hyperparameter, the information bottleneck $\epsilon$, while HarmonyDream does not introduce any hyperparameters.

\subsection{Learned Coefficients}
\begin{figure*}[h]
    \centering
    \begin{minipage}[t]{0.29\textwidth}
        \centering
        \includegraphics[width=\textwidth]{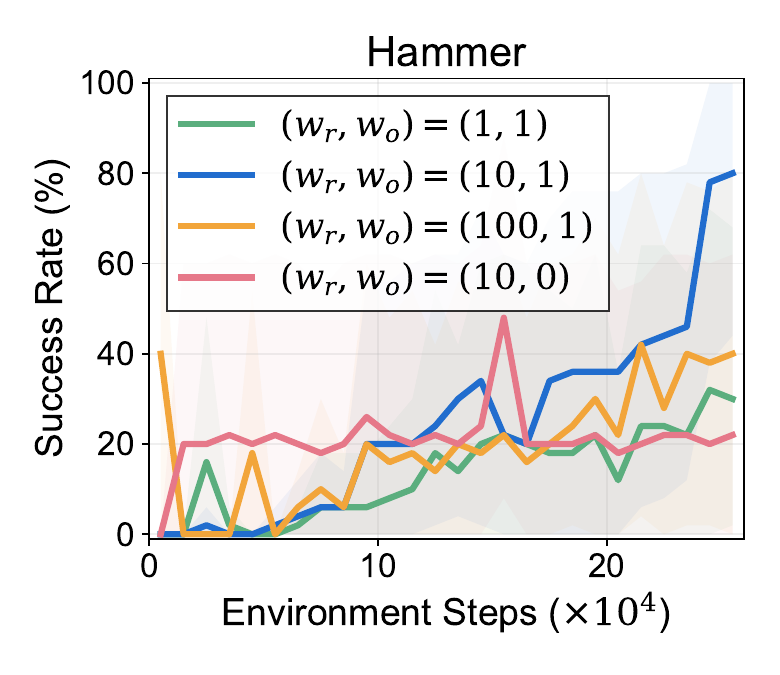}
        {\vspace{-12pt}
        \caption{Effects of different loss coefficients on an additional task.}
        \label{fig:reward_coeff_hammer}}
    \end{minipage}
    \hfil
    \begin{minipage}[t]{0.55\textwidth}
        \centering
        \includegraphics[width=\textwidth]{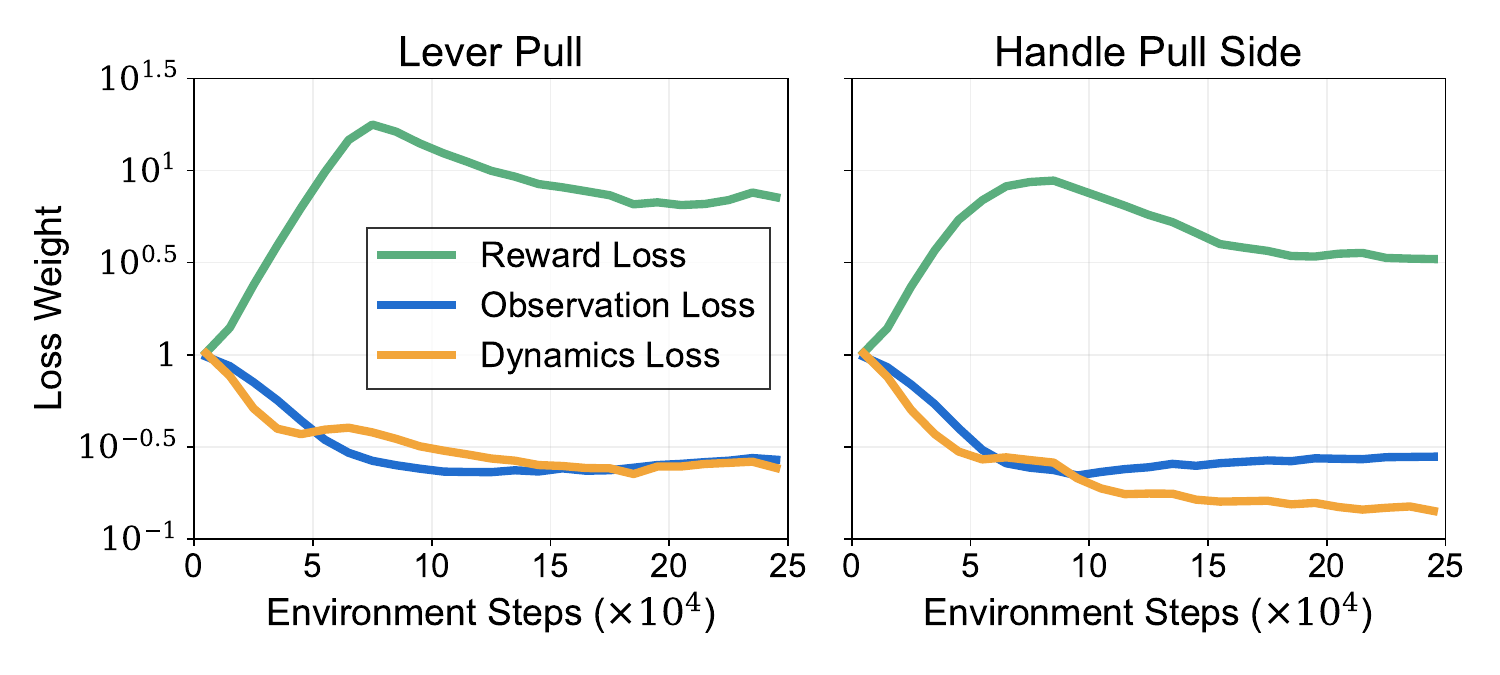}
        {\vspace{-12pt}
        \caption{Learned harmony loss coefficients on Meta-world tasks.}
        \label{fig:learned_uncertainty_coeff}}
    \end{minipage}
\end{figure*}

Fig.~\ref{fig:learned_uncertainty_coeff} illustrates the learned harmony loss coefficients for two Meta-world tasks: Lever Pull and Handle Pull Side. The harmonized reward loss coefficient for Lever Pull is observed to be higher than that for Handle Pull Side. This observation aligns with the fact that the coefficient pair $(w_r,w_o)=(100,1)$ yields superior performance on Lever Pull, while the pair $(w_r,w_o)=(10,1)$ facilitates better learning on Handle Pull Side, as depicted in Fig.~\ref{fig:effect_coeff_metaworld}. 

Additionally, we present the impact of varying loss coefficients for DreamerV2 on the Meta-world Hammer task in Fig.~\ref{fig:reward_coeff_hammer}, supplementing the information in Fig.~\ref{fig:effect_coeff_metaworld}.

\subsection{Quantitative Evaluation of the Beneficial Impact of Observation Modeling on Reward Modeling}
To explore the possible beneficial impact of observation modeling on reward modeling, we utilize the offline experimental setup in Fig~\ref{fig:representation_difference} and \ref{fig:r1_r100_video_prediction_mw}, whose details are described in Appendix \ref{app:qualitative}.
We offline train two DreamerV2 agents using task weights $(w_r=100, w_o=1)$ and $(w_r=100, w_o=0)$ and evaluate the ability to accurately predict rewards on a validation set with the same distribution as the offline training set. For this task, we gathered 20,000 segments of trajectories, each of length 50. We utilized 35 frames for observation and predicted the reward for the remaining 15 frames. Results are reported in the form of average MSE loss. We observe that the world model with observation modeling predicts the reward better than the world model that only models the reward. The prediction loss of $(w_r=100, w_o=1)$ is $0.379$, while the loss of $(w_r=100, w_o=0)$ is $0.416$. This result indicates that observation modeling has a positive effect on reward modeling.

\end{document}